\documentclass{article}

\PassOptionsToPackage{numbers, compress}{natbib}

\usepackage[preprint]{neurips_2026}

\usepackage[utf8]{inputenc} % allow utf-8 input
\usepackage[T1]{fontenc}    % use 8-bit T1 fonts
\usepackage{hyperref}       % hyperlinks
\usepackage{url}            % simple URL typesetting
\usepackage{booktabs}       % professional-quality tables
\usepackage{amsfonts}       % blackboard math symbols
\usepackage{nicefrac}       % compact symbols for 1/2, etc.
\usepackage{microtype}      % microtypography
\usepackage{xcolor}         % colors
\usepackage{graphicx}
\usepackage{algorithm}
\usepackage{algpseudocode}
\usepackage{amsmath}
\usepackage{caption} % Check to make sure it doesn't mess up anything
\usepackage{tikz}
\usepackage{xspace}
    \title{Spatially-Grounded Flow Matching: Structured Source Distributions for Image Generation}

\author{Arman Zarei\textsuperscript{1,2}\thanks{Work done during internship at Netflix} , Mahdi M. Kalayeh\textsuperscript{2}\\
  \textsuperscript{1}University of Maryland \hspace{5pt} \textsuperscript{2}Netflix}

\newcommand{\Input}{\item[\textbf{Input:}]}
\newcommand{\Output}{\item[\textbf{Output:}]}
\newcommand{\modelname}{StructFlow\xspace}
\begin{document}
\maketitle

\vspace{-8pt}
\begin{abstract}
\vspace{-4pt}
Current flow matching models learn to transport the source i.i.d. Gaussian noise into the target distribution of natural images, yet this source distribution carries no notion of spatial structure. Images however are fundamentally local since nearby pixels are strongly correlated. By sampling the noise independently, we hypothesize that models are implicitly encouraged to exploit less noisy neighbors as context during training, partially bypassing the need to properly learn the true local structure of images. The source distribution, in other words, works against the inductive bias of the image domain. To ameliorate this design discrepancy, we propose \modelname which encodes spatial locality directly into the source by having the pixels within a small region share a common noise component. This structured source produces transport paths that are geometrically aligned with image regions — enabling properties that generic flow matching struggles to provide: fine-grained local editing that naturally respects boundaries, robust structure preservation, and smooth semantic interpolation between images. We show that these benefits also extend to large pre-trained models, demonstrating that \modelname can even be incorporated through a lightweight post-training phase. Comprehensive experiments on multiple datasets, in unconditional, class and text-conditioned regimes, using different diffusion transformer architectures confirm that \modelname not only offers competitive image generation quality, but also significantly improves localized controllable re-synthesis. \footnote[1]{Project page is available at: \href{https://armanzarei.github.io/StructFlow}{https://armanzarei.github.io/StructFlow}}

\end{abstract}
\section{Introduction}

Recent advances in image generative modeling, particularly diffusion models and flow matching~\cite{ho2020denoising, rombach2022high, nichol2021improved, ho2022classifier, flux2024, esser2024scaling, lipman2022flow, zarei2024improving, zarei2024understanding, wu2025qwen}, have led to remarkable improvements in visual quality and scalability, establishing state-of-the-art across a wide range of tasks~\cite{ruiz2023dreambooth, brooks2023instructpix2pix, zarei2026slideredit, sohn2023styledrop, zarei2026localizing, wang2024instantstyle, zarei2025agentcomp, zhang2023adding}. In flow matching, generation is formulated as learning a continuous transport from a simple source distribution—typically an isotropic Gaussian—to the data distribution via a learned velocity field. Despite its success, this standard choice of source distribution is largely inherited from diffusion models and remains underexplored. A key limitation of this formulation is a fundamental mismatch between the source distribution and that of natural images. While images exhibit strong spatial locality—nearby pixels are highly correlated and form coherent regions—the commonly used i.i.d. Gaussian prior assumes complete independence across spatial locations. As a result, the model is implicitly required to reconstruct this locality through its learned dynamics, rather than benefiting from it as an inductive bias. This places an unnecessary burden which we hypothesize would hinder the training efficiency as well as controllability of the final model. Several recent works have explored alternatives to the standard Gaussian source distribution, including non-Gaussian priors~\cite{nachmani2021non, xu2022poisson}, mixture-based formulations~\cite{chen2025gaussian, jia2024structured}, and learned source distributions~\cite{kim2026better, lee2025there}. While these approaches improve aspects such as convergence and expressivity, they largely remain agnostic to the spatial structure inherent in the natural images. In contrast, structured noise has proven highly effective in other domains: for instance, recent video diffusion models introduce temporally correlated noise to enforce consistency across frames~\cite{ge2023preserve, chang2025warped}, leading to significantly improved coherence. This raises a natural question: \emph{can similar structured priors be leveraged in the spatial domain for image generation?}

In this work, we introduce \emph{\modelname}, a flow matching framework with spatially-grounded source distributions that explicitly encodes locality. Our key idea is to construct a structured prior in which pixels within a small region share a common noise component, inducing a bias that aligns with local structure in natural images. This simple modification leads to transport paths that are geometrically consistent with image regions, enabling emergent capabilities such as fine-grained local editing, improved structure preservation, and controllable coherence during generation. 
However, a naive implementation of such a structured source distribution introduces optimization challenges, mainly due to increased correlations in the noise, which severely destabilize the training. To facilitate the learning process, we propose (i) a cosine-induced spatial anchoring which reduces stochastic variability while providing consistent spatial cues, (ii) a progressive coherence annealing schedule that gradually introduces structure during training, and (iii) a mixed-coherence training strategy that enables a single model to operate across a spectrum that trades generation quality for localized controllable re-synthesis capabilities.
We validate \modelname across multiple problem settings, including class and text-conditioned image generation, demonstrating that our approach achieves competitive image quality while significantly improving the granularity of control when it comes to local edits (\textit{i.e.} re-synthesis). Notably, our method can also be incorporated into pre-trained models through a lightweight post-training, making it applicable to existing foundation generative models.
% \todo{we should refer to how patch size has been shown crucial in what MAEs learn i.e sufficiently large patches are needed otherwise models cheat by looking at unmasked neighbors. same analogy goes with low masking ratios}

In summary, our contributions are:
(1) We identify the mismatch between i.i.d. source distributions and spatial locality in images as a key limitation of standard flow matching and propose \emph{\modelname}, a spatially structured source distribution that encodes locality directly into the generative process.
(2) We introduce a set of training strategies that stabilize optimization under structured noise while preserving its benefits.
(3) We demonstrate improved controllability, structure preservation, and editing capabilities without sacrificing generation quality across multiple setups and benchmarks.

\section{Background}

Our work builds on flow matching~\citep{lipman2022flow, liu2022flow}, which learns a time-dependent vector field to transport samples from a source distribution to the data distribution via an ODE (see Appendix~\ref{app:related_works_diffusion_and_flow} for details).
A key design choice in flow matching is the source distribution, which is typically taken to be a standard isotropic Gaussian—largely inherited from diffusion models rather than optimized for the task. A growing body of work challenges this assumption. For instance, Cold Diffusion~\citep{bansal2023cold} shows that arbitrary image degradations can define valid generative processes; Non-Gaussian Diffusion~\citep{nachmani2021non} and Poisson Flow~\citep{xu2022poisson} explore alternative priors such as Gamma distributions and hemisphere-uniform sampling; and Gaussian Mixture Flow Matching~\citep{chen2025gaussian} and Structured Diffusion with MoG~\citep{jia2024structured} employ mixture-based priors for richer distributional coverage. More directly related to our setting, several works investigate the design of source distributions in flow matching~\citep{kim2026better, lee2025there, zwick2025lediflow}, demonstrating that carefully chosen or learned priors can significantly improve convergence speed and generation quality. Whitened Score Diffusion~\citep{alido2025whitened} further highlights the benefits of anisotropic Gaussian priors by introducing spectral inductive biases tailored to imaging tasks. Additional approaches construct non-standard priors via normalizing flows~\citep{zand2023diffusion} or diffusion bridges~\citep{zhou2023denoising}, or introduce spatially varying noise through time-dependent scheduling on pre-trained models~\citep{yu2023constructing}. 
% \mk{any of these deserve to be compared against ours? if so, why have not we done it. if not, we should make it clear so reviewer do not expect them in the experimental results} 
Beyond images, recent work has shown that introducing structured correlations into the noise prior, rather than treating pixels independently, can improve coherence in video generation. For example, PYoCo~\citep{ge2023preserve} proposes a mixed noise prior for video diffusion in which frames share a common noise component, enabling strong temporal consistency. Similarly, \citep{chang2025warped} introduces a principled method to warp noise across video frames using optical flow, preserving noise statistics while enforcing temporal alignment.

Our work complements and extends these directions by introducing a spatially structured source distribution that explicitly captures the locality inherent in natural images. We argue that this inductive bias should be built into the source distribution itself, rather than learned entirely from data through an unstructured Gaussian prior. Concretely, we construct priors in which pixels within the same spatial neighborhood share correlated noise, thereby promoting spatial coherence throughout the flow matching process. As a result, the model naturally learns to preserve structural consistency during generation, while enabling emergent capabilities such as localized editing. 
% Removed "and respect object boundaries"

\section{\modelname: Structured Source Distributions for Flow Matching}

\begin{figure}[t]
\centering
\begin{tikzpicture}
    \node[anchor=south west, inner sep=0] (main) at (0,0) {
        \includegraphics[width=\linewidth]{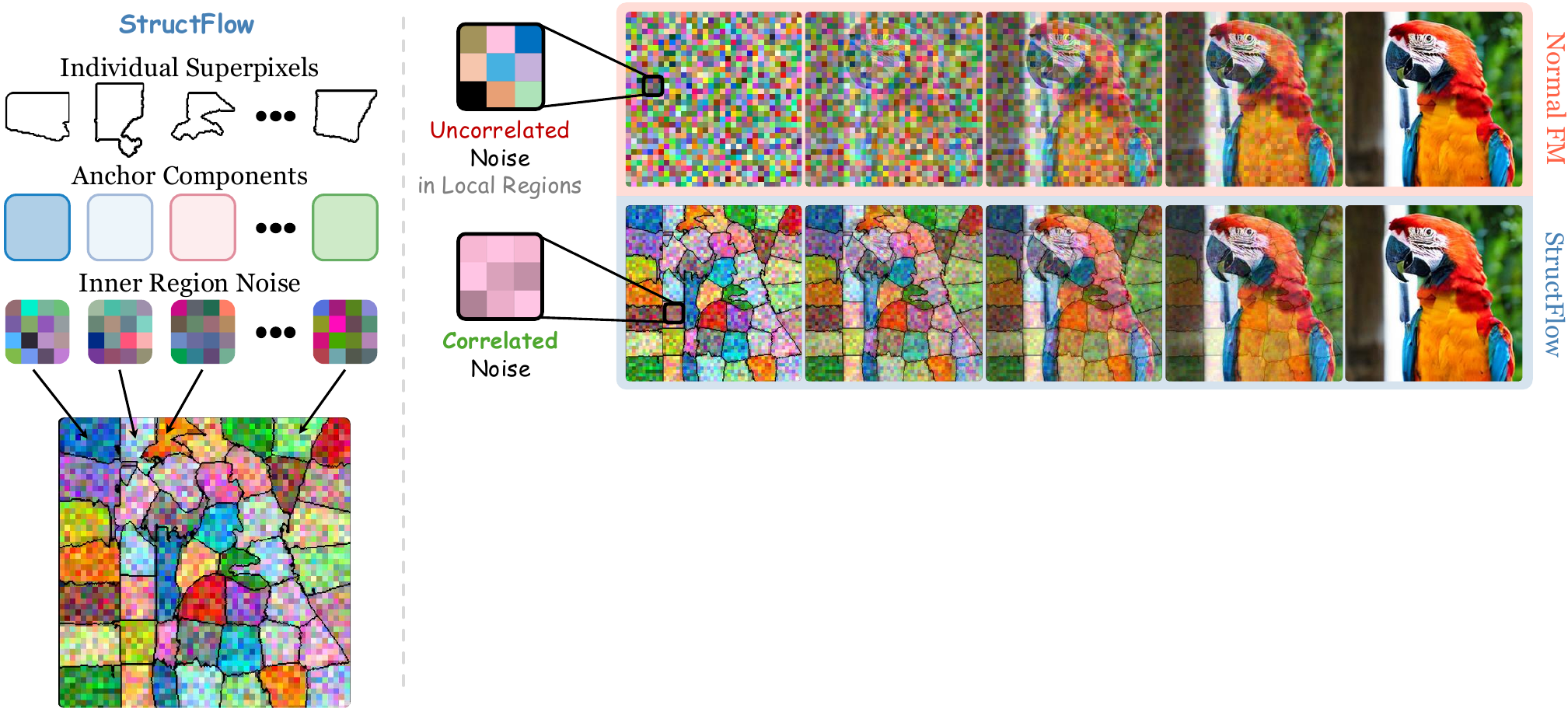}
    };
    \node[anchor=south east, inner sep=0, xshift=-8mm] at (main.south east) {
        \includegraphics[width=0.63\linewidth]{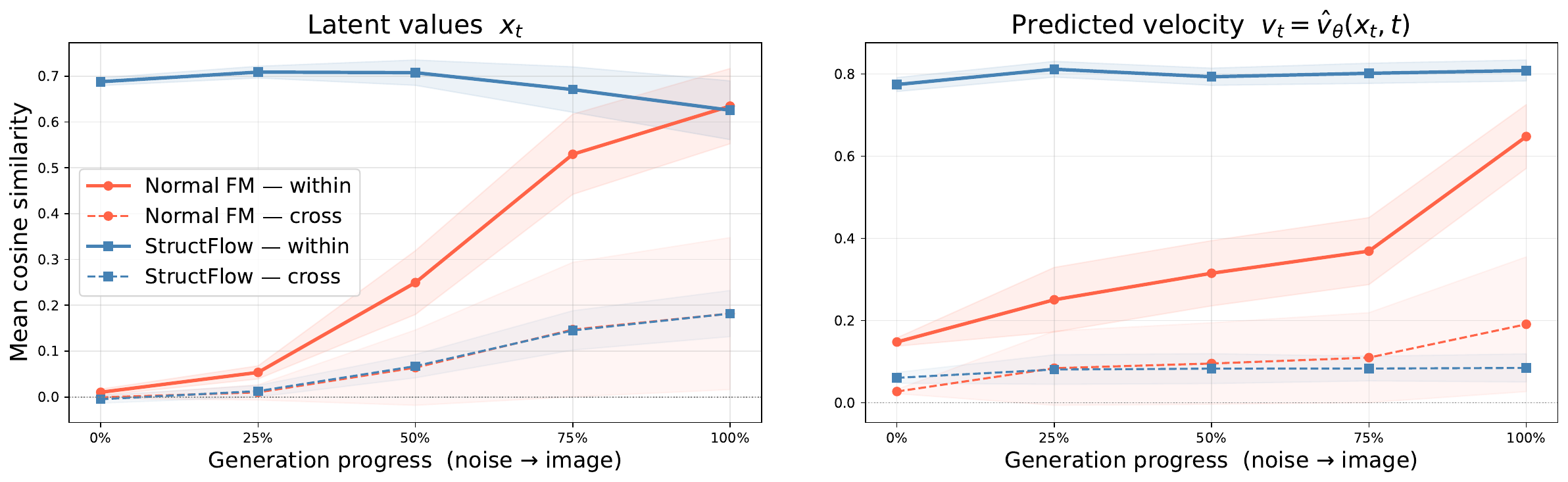}
    };
\end{tikzpicture}
\caption{\textbf{\modelname Noise Construction and locality analysis.} \modelname introduces locality in the source distribution instead of i.i.d gaussian sampling in normal flow matching. Locality emerges implicitly in standard flow matching, while \modelname enforces it directly in the source distribution.}
\label{fig:noise_construction}
\label{fig:interpretability}
\end{figure}

%% ─────────────────────────────────────────────────────────────────────────────
\subsection{Flow Matching and Locality Bias}
%% ─────────────────────────────────────────────────────────────────────────────

As a starting point, we briefly review the standard flow matching~\cite{lipman2022flow} framework, in particular rectified flow~\cite{liu2022flow}, upon which our method is built. Flow matching constructs a continuous-time transport map between a source distribution $p_0$ and the data distribution $p_1$. In practice, for image generation, the source distribution is typically chosen as an isotropic Gaussian, $p_0 = \mathcal{N}(\mathbf{0}, \mathbf{I})$, such that each pixel (or latent dimension) is sampled independently. Given a data sample $\mathbf{x}_1$ and a source sample $\boldsymbol{\epsilon} \sim p_0$, rectified flow considers the linear interpolation path
\begin{equation}
  \mathbf{x}_t = (1 - t)\,\boldsymbol{\epsilon} + t\,\mathbf{x}_1, \qquad t \in [0, 1],
\end{equation}
% $\mathbf{x}_t = (1 - t)\,\boldsymbol{\epsilon} + t\,\mathbf{x}_1$ ($t \in [0, 1]$),
and learns a neural velocity field $v_\theta(\mathbf{x}_t, t)$ to predict the corresponding constant velocity, $\mathbf{x}_1 - \boldsymbol{\epsilon}$.

% \paragraph{Mismatch between i.i.d.\ noise and image locality inductive bias.}
% \mk{no need to create bold paragraph headers, you lose space and the subsection is already short}
Natural images exhibit a strong \emph{locality inductive bias}: nearby pixels are highly correlated and form coherent spatial regions. In contrast, the standard choice of source distribution, $p_0 = \mathcal{N}(\mathbf{0}, \mathbf{I})$, is fully factorized, producing spatially independent noise. As a result, the model must reconstruct this locality entirely through the learned velocity field, forcing the model to reconstruct spatial structure through the learned velocity field, rather than inheriting it from the source distribution.

% \paragraph{Empirical evidence.}
To study this effect, we analyze a DiT-based flow matching model. We sample generated trajectories and record both the latent state $\mathbf{x}^i_t$ and predicted velocity $v_\theta(\mathbf{x}^i_t, t)$ at multiple timesteps $t \in \{0, 0.25, 0.5, 0.75, 1\}$. For each generated sample $\mathbf{x}^i_1$, we compute a SLIC~\cite{achanta2012slic} superpixel segmentation $\mathbf{M}^i$ and measure feature similarity within and across segments in both $\mathbf{x}^i_t$ and $v_\theta(\mathbf{x}^i_t, t)$.
% \mk{remind the reader that SLICE runs on final generated images not noisy intermediate versions BUT we then overlap it on intermediate representations. Actually, here making figure with real data would significantly help. also, I think fig2 should actually be situated here as a wide yet short figure showing all the noisy versions of the image}
Let $\mathbf{f} \in \{\mathbf{x}_t,\; v_\theta(\mathbf{x}_t, t)\} \in \mathbb{R}^{N_p \times C}$ denote the pixel features (flattened spatial dimensions)
and $\hat{\mathbf{f}}_p = \mathbf{f}_p / \|\mathbf{f}_p\|$ the $\ell_2$-normalised features. Given a segmentation $\mathbf{M}^i$ with segments $\{\mathcal{S}_k\}_{k=1}^K$ ($\mathcal{S}_k$ denotes the set of pixels in $k$-th segment), we define the average within-segment and cross-segment cosine similarities:
\begin{align}
  \rho_{\mathrm{in}}(\mathbf{f}, \mathbf{M})  &= \mathbb{E}_{k}\;\mathbb{E}_{p,q \in \mathcal{S}_k}\left[\hat{\mathbf{f}}_p^\top \hat{\mathbf{f}}_q\right] = \frac{1}{K}\sum_{k=1}^{K}
    \frac{1}{\binom{|\mathcal{S}_k|}{2}} \sum_{\substack{p,q \in \mathcal{S}_k \\ p < q}}
    \hat{\mathbf{f}}_p^\top \hat{\mathbf{f}}_q, \\
  \rho_{\mathrm{out}}(\mathbf{f}, \mathbf{M}) &= \mathbb{E}_{k}\;\mathbb{E}_{p \in \mathcal{S}_k,\, q \notin \mathcal{S}_k}\left[\hat{\mathbf{f}}_p^\top \hat{\mathbf{f}}_q\right] = \frac{1}{K}\sum_{k=1}^{K}
    \frac{1}{|\mathcal{S}_k||\bar{\mathcal{S}}_k|}
    \sum_{p \in \mathcal{S}_k,\, q \in \bar{\mathcal{S}}_k}
    \hat{\mathbf{f}}_p^\top \hat{\mathbf{f}}_q.
\end{align}
% where $\mathcal{S}_k$ denotes the set of pixels in segment $k$ and $\bar{\mathcal{S}}_k$ its complement.

\paragraph{Findings.}
As shown in Fig.~\ref{fig:interpretability}, the latent state $\mathbf{x}_t$ exhibits almost no spatial structure at initialization ($\rho_{\mathrm{in}}(x_t, M) \approx 0$ at $t{=}0$), and gradually acquires locality over time. In contrast, the velocity field already encodes strong spatial correlations from the earliest steps: even at $t{=}0$, we observe a clear gap $\rho_{\mathrm{in}} - \rho_{\mathrm{out}}$, which increases steadily during generation. This indicates that the model learns to impose spatial structure through its velocity field, compensating for the lack of structure in the source distribution. In other words, part of the model capacity is spent reconstructing locality that could instead be built into $p_0$.
% \mk{you should stress the different pattern of SF vs FM before proceeding any further.}
This observation motivates our central question: \emph{can we design a source distribution $p_0$ that already respects
the locality structure of images, relieving the model of this burden
and unlocking emergent capabilities?}

\begin{figure}[t]
\begin{minipage}[t]{0.50\textwidth}

\begin{algorithm}[H]
  \caption{\small \modelname Hierarchical Noise Sampling}
  \label{alg:seg_aware_noise}
  \begin{algorithmic}[1]
  \small
  \Input {\footnotesize Mask $\mathbf{M} \in \mathbb{Z}_{\geq 0}^{H \times W}$ with segments $\{\mathcal{S}_k\}_{k=1}^K$,
           latent spatial size $(C, H, W)$,
           secondary noise scale $\lambda_\varsigma \geq 0$,
           cosine grid $g \in [-1, 1]^{H\times W}$}
  \Output {\footnotesize Structured noise $\boldsymbol{\epsilon} \in \mathbb{R}^{C \times H \times W}$}

  \vspace{0.4em}
  \State \textit{\footnotesize \textcolor{gray}{\# \textbf{1st Level}: Sample one shared noise vector per segment}}
  \For{$k = 1, \ldots, K$}
    % \If{{\footnotesize $\texttt{Cosine\_Anchoring}$}}
    %     \State $\mathbf{z}_k \leftarrow|\mathcal{S}_k|^{-1}\sum_{(i,j) \in \mathcal{S}_k} g(i, j)$
    % \Else
    %     \State $\mathbf{z}_k \sim \mathcal{N}(\mathbf{0},\, \mathbf{I}_C)$
    % \EndIf
    \State $\mathbf{z}_k\leftarrow\begin{cases}
    |\mathcal{S}_k|^{-1}\sum_{(i,j)\in\mathcal{S}_k} g(i,j), & \text{if \texttt{CISA}},\\
    \mathcal{N}(\mathbf{0},\mathbf{I}_C), & \text{otherwise}.
    \end{cases}$
  \EndFor

  \vspace{0.4em}
  \State \textit{\textcolor{gray}{\footnotesize \# \textbf{2nd Level}: Sample per-pixel noise per shared segment}}
  \For{each spatial location $(i, j)$}
      \State $\boldsymbol{\epsilon}_{i,j} \sim \mathcal{N}(\mathbf{z}_{\mathbf{M}(i,j)},\, \lambda_\varsigma^2 \mathbf{I}_C)$
  \EndFor

  \State \Return $\boldsymbol{\epsilon} / \sqrt{1 + \lambda_\varsigma^2}$ \Comment{\textcolor{gray}{\footnotesize Rescale to unit variance}}
  \end{algorithmic}
\end{algorithm}

\end{minipage}
\hfill
\begin{minipage}[t]{0.48\textwidth}
\begin{algorithm}[H]
  \caption{\small \modelname Training}
  \label{alg:structflow_training}
  \begin{algorithmic}[1]
  \small
  \Input {\small Dataset $\mathcal{D}$, Model $v_\theta$, \# of Superpixels $K$}
  \For{each training step $s$}
      \State Sample $\mathbf{x}_1 \sim \mathcal{D}$
      \State $\mathbf{M} = \texttt{SLIC}(\mathbf{x}_1, K)$ 
      \If{{\footnotesize $\texttt{Progressive\_Coherence\_Annealing}$}}
          \State Update {\footnotesize$\lambda_\varsigma \leftarrow  \text{schdeuler}_{\lambda_\varsigma}(s, \lambda_{\min}, \lambda_{\max})$}
      \EndIf
      \If{{\footnotesize $\texttt{Mixed\_Coherence} \;\&\; \text{schdeuler}_{\lambda_\varsigma}.\text{finish}$}}
          \State Sample $\lambda_\varsigma \sim \mathcal{U}[\lambda_{\min}-\delta,\lambda_{\min}+\delta]$
      \EndIf
      \State Sample $\boldsymbol{\epsilon}$ using Alg.~\ref{alg:seg_aware_noise} and $t \sim \mathcal{U}(0,1)$
      \State $\mathbf{x}_t \leftarrow (1-t)\boldsymbol{\epsilon} + t\mathbf{x}_1$
      \State $\hat{\mathbf{v}}_\theta \leftarrow v_\theta(\mathbf{x}_t, t)$
      \State $\mathcal{L} \leftarrow \|\hat{\mathbf{v}}_\theta - (\mathbf{x}_1 - \boldsymbol{\epsilon})\|_2^2$
      \State Update $\theta$ with $\nabla_\theta \mathcal{L}$
  \EndFor
  \end{algorithmic}
\end{algorithm}

\end{minipage}

\end{figure}

%% ─────────────────────────────────────────────────────────────────────────────
\subsection{Structured Hierarchical Noise Sampling}
\label{sec:method:base_noise_sampling}
%% ─────────────────────────────────────────────────────────────────────────────

We introduce \textbf{\modelname}, which replaces the isotropic Gaussian source with a
\emph{structured} distribution that encodes spatial locality. The key idea is to induce
correlations among pixels within the same region, while preserving sufficient stochasticity
to maintain full support.

% \paragraph{Setup.}
Let $\mathbf{M} \in \mathbb{Z}_{\geq 0}^{H \times W}$ denote a superpixel decomposition with $K$ regions,
obtained via SLIC oversegmentation~\cite{achanta2012slic} (or a proxy at inference; see
Section~\ref{sec:mask_generation}). We operate in the VAE latent space of shape $(C, H, W)$.
% \paragraph{Structured source distribution.}
We define a hierarchical Gaussian prior in which pixels within the same segment share a
common latent anchor. Concretely, for each segment $k \in \{1, \ldots, K\}$, we sample an anchor vector
$\mathbf{z}_k \sim \mathcal{N}(\mathbf{0}, \mathbf{I}_C)$, and for each spatial location
$(i,j)$ belonging to segment $k = \mathbf{M}(i,j)$, we sample 
% \mk{cannot we combine them in one line where both Normals are zero-mean? then you no longer need to explicitly define $z_{k}$}
\begin{equation}
  \boldsymbol{\epsilon}_{i,j} \sim \mathcal{N}\!\left(\mathbf{z}_k,\; \lambda_\varsigma^2\,\mathbf{I}_C\right),
\end{equation}
followed by a global rescaling $\boldsymbol{\epsilon} \;\leftarrow\; \frac{\boldsymbol{\epsilon}}{\sqrt{1 + \lambda_\varsigma^2}}$ to ensure unit marginal variance.
Full sampling procedure is provided in Alg.~\ref{alg:seg_aware_noise} and in Fig~\ref{fig:noise_construction}. 
This construction induces a simple correlation structure: pixels within the same local region
have correlation 
\begin{equation}
  \mathrm{Corr}(\epsilon_{i,j},\, \epsilon_{i',j'}) = \frac{1}{1 + \lambda_\varsigma^2}
  \quad \text{for } \mathbf{M}(i,j) = \mathbf{M}(i',j'),
\end{equation}
while pixels from different segments
remain uncorrelated. The parameter $\lambda_\varsigma$ therefore directly controls the
strength of locality, smoothly interpolating between fully correlated segments
($\lambda_\varsigma \to 0$) and standard i.i.d.\ Gaussian noise ($\lambda_\varsigma \to \infty$) (See Fig~\ref{fig:secondary_std_to_inf_visualization} in Appendix).

% \paragraph{Training and inference.}
Aside from the structured source $\boldsymbol{\epsilon}$, we make no modifications
to the flow matching training objective or inference procedure.
The model is trained to predict $\hat{v}_\theta(\mathbf{x}_t, t)$ with the standard MSE loss,
and DDIM/Euler integration is used at inference — the only change being that the initial
noise is drawn from our structured distribution rather than $\mathcal{N}(\mathbf{0}, \mathbf{I})$.

% \paragraph{Segmentation flexibility.}
% The method does not rely on precise semantic segmentation. While we use SLIC during training,
% simpler alternatives at inference—such as regular grids or approximate segmentations—are
% sufficient to induce meaningful locality (see Section~\ref{sec:experiments}).
% \paragraph{Generality of the segmentation source.}
% While we primarily use SLIC superpixels computed from the training image,
% the source distribution does not require a semantically precise segmentation.
% At inference, simple alternatives such as regular grid patches or outputs
% from lightweight generative models also yield strong locality — we study this in Section~\ref{sec:experiments}.

% \paragraph{Emergent properties.}
% Injecting locality into the source distribution enables several capabilities without
% architectural changes: localized editing via segment-wise perturbations, improved structural
% coherence, and controllable generation behavior through $\lambda_\varsigma$.

% \paragraph{Emergent capabilities.}
Encoding locality in the source distribution, as reflected in the analysis in Fig.~\ref{fig:interpretability}, makes local regions more coherent and better separated from the rest of the image throughout generation. This gives rise to several downstream properties \emph{out of the box}, including \emph{local region editing} (Sec.~\ref{sec:part_resampling}), \emph{structure-preserving generation} (Sec.~\ref{sec:structure_preserving_generation}), and the \emph{early emergence of informative features} (Sec.~\ref{sec:informative_features}).
% \mk{similar to previous comment, you will lose space by creating bold paragraph headers and for such a short text block it is not even justified}

%% ─────────────────────────────────────────────────────────────────────────────
\subsection{StructFlow Training Strategies}
\label{sec:training_strategies}
%% ─────────────────────────────────────────────────────────────────────────────

Naïvely training a DiT on the structured source distribution described above
can lead to optimization instability: the induced correlations increase gradient variance, making it difficult
for the model to learn a consistent velocity field from highly correlated noise.
We identify three complementary strategies that resolve this instability
while preserving, and enhancing, the locality capabilities.

\subsubsection{Cosine-Induced Spatial Anchoring}

% In the basic formulation (Algo.~\ref{alg:seg_aware_noise}),
% the segment anchor $\mathbf{z}_k$ is sampled independently at each training step, making the per-segment noise values a highly variable random quantity.
% \mk{the current wording could imply that the issue is with changing $z_{k}$ from one epoch to another but IIUC the more crucial problem is that even within the same step, two neighboring blob could have had very different anchors hence making the training difficult. this aspect is not delivered here}
In the basic formulation (Alg.~\ref{alg:seg_aware_noise}), each segment anchor
$\mathbf{z}_k$ is sampled independently from a Gaussian distribution. Although this
creates within-segment coherence, it can also assign very different anchors to adjacent
segments within the same training sample, introducing abrupt high-variance
discontinuities in the structured source that makes the target velocity field harder to
learn.
We find that replacing this random sampling with a \emph{reduced-stochasticity spatial structure}
substantially stabilises training. Concretely, we define a diagonal cosine grid over the spatial lattice:
\begin{equation}
  g(i, j) = 2 \cdot \frac{1 - \cos\!\left(\pi \cdot \tfrac{i + j}{2(H-1)}\right)}{2} - 1
  \;\in [-1, 1], \quad i \in [0, H{-}1],\; j \in [0, W{-}1].
\end{equation}
Each segment $k$ is assigned the mean grid value over its pixels,
$\bar{g}_k = |\mathcal{S}_k|^{-1}\sum_{(i,j) \in \mathcal{S}_k} g(i, j)$,
and the first-level anchor is replaced by the scalar broadcast
$\mathbf{z}_k \;\leftarrow\; \bar{g}_k \cdot \mathbf{1}_C$ (See Fig~\ref{fig:appendix_cosine_anchor} in Appendix).
% The second-level perturbation is then scaled relative to the spread of segment values:
% \begin{equation}
%   \boldsymbol{\epsilon}_{i,j}
%     = \bar{g}_{\mathbf{M}(i,j)}\cdot\mathbf{1}_C
%     + \lambda_\varsigma \cdot \sigma_{\bar{g}} \cdot \boldsymbol{\xi}_{i,j},
%   \quad \boldsymbol{\xi}_{i,j} \sim \mathcal{N}(\mathbf{0}, \mathbf{I}_C),
% \end{equation}
% where $\sigma_{\bar{g}} = \mathrm{std}_k(\bar{g}_k)$ normalises the secondary noise
% to the scale of the primary structure.
% The combined noise is then rescaled to unit per-sample standard deviation. \az{TODO: remove the extra details such as the normalization and ...}
% \mk{include an image example of the grid here to better deliver the intuition}
This Cosine-Induced Spatial Anchoring (CISA) provides two benefits:
(i) the first-level structure is deterministic given the mask,
removing one source of stochastic variation from the training signal;
(ii) the smooth diagonal gradient gives the model a consistent spatial cue which it can leverage throughout training. Overall, this yields a more well-conditioned learning problem, leading to improved optimization stability as shown in Sec~\ref{sec:experiments:qualitative_and_quantitative} and Table~\ref{tab:training_ablation}.

\subsubsection{Progressive Coherence Annealing}
\label{sec:method:progressive_annealing}

During our experiments, we realized that the noise scale $\lambda_\varsigma$ governs a fundamental
\emph{quality–editability trade-off}: small $\lambda_\varsigma$ yields
tightly coherent segments that enable precise local editing,
but makes the source distribution far from Gaussian, destabilizing training.
Large $\lambda_\varsigma$ approaches the i.i.d.\ baseline, easing optimisation
at the cost of weaker locality guarantees.
% \mk{should we use some simpler symbol like $\gamma$?}

We resolve this tension with a \emph{coherence annealing} schedule:
$\lambda_\varsigma$ is initialised to a large value $\lambda_{\max}$
and linearly decreased to a target $\lambda_{\min}$ over an interval of training steps $[s_0, s_1]$: 
% \mk{have we defined what $s_0$ and $s_1$ are?}
\begin{equation}
  \lambda_\varsigma(s) = \lambda_{\max} - (\lambda_{\max} - \lambda_{\min}) \cdot
  \frac{\min(s, s_1) - s_0}{s_1 - s_0}, \quad s \geq s_0.
\end{equation}
This curriculum first trains the model on a near-Gaussian source (easy),
then progressively tightens the locality constraint as the model matures.

\paragraph{Post-training a standard FM model.}
Coherence annealing enables a particularly practical workflow:
a standard flow matching model can be
\emph{fine-tuned} into a \modelname model by applying the annealing schedule
starting from the pretrained checkpoint.
Because the initial $\lambda_\varsigma = \lambda_{\max} \rightarrow \infty$ is close to the training distribution
of the base model, fine-tuning is stable; the model gradually adapts to the
structured source without catastrophic forgetting.
We demonstrate this in Sec.~\ref{sec:experiments:t2i}.

\subsubsection{Mixed-Coherence Training}

Even with annealing, committing to a single final $\lambda_\varsigma$ at test time
forces a binary choice between generation quality (high $\lambda_\varsigma$)
and locality control (low $\lambda_\varsigma$).
We instead propose \emph{mixed-coherence training}: As the annealing schedule approaches $\lambda_{\min}$, at each training iteration,
$\lambda_\varsigma$ is sampled uniformly from a range $[\lambda_{\min}-\delta, \lambda_{\min}+\delta]$.
This provides two advantages.
First, training across a distribution of coherence levels acts as a regulariser,
improving generation quality even at the hard end ($\lambda_\varsigma = \lambda_{\min}-\delta$)
compared to training at that value alone.
Second, a single trained model can serve the full spectrum of use cases at inference:
low $\lambda_\varsigma$ for fine-grained local editing and structure preservation,
high $\lambda_\varsigma$ for maximum image quality with mild locality bias.
Users can thus navigate the quality–editability Pareto frontier \emph{without retraining}. Algorithms~\ref{alg:seg_aware_noise} and \ref{alg:structflow_training} summarize our stabilization techniques for training \modelname. 
% \mk{I think having a figure illustrating how $\lambda_\varsigma$ is evolving is very helpful}

\section{Experiments}

In this section, we conduct comprehensive quantitative and qualitative evaluations of \modelname, demonstrating strong performance across a wide range of metrics while revealing compelling emerging capabilities that arise naturally from our approach.

\subsection{Implementation details}
We use DiT-XL~\cite{peebles2023scalable} as the base architecture in all experiments, building a flow matching implementation similar to SiT~\cite{ma2024sit}. Unless otherwise stated, all models are trained from scratch. We use a global batch size of 128 for class-conditional ImageNet and 64 for unconditional FFHQ. For class-conditional training, we apply class dropout with probability 0.1 to enable classifier-free guidance (CFG) at inference, where we use a fixed guidance scale of 4.0 across all evaluations. We train on FFHQ for 130k iterations and on ImageNet for approximately 3M iterations unless mentioned otherwise.
% \mk{perhaps unless mentioned otherwise. IIRC some of the ablations were not trained for 3M}
Additional implementation details are provided in Appendix~\ref{sec:app:experiments:implementation_details}.

\subsection{Qualitative and Quantitative evaluation}
\label{sec:experiments:qualitative_and_quantitative}

\begin{figure}[t]
    \centering
    \includegraphics[width=\linewidth]{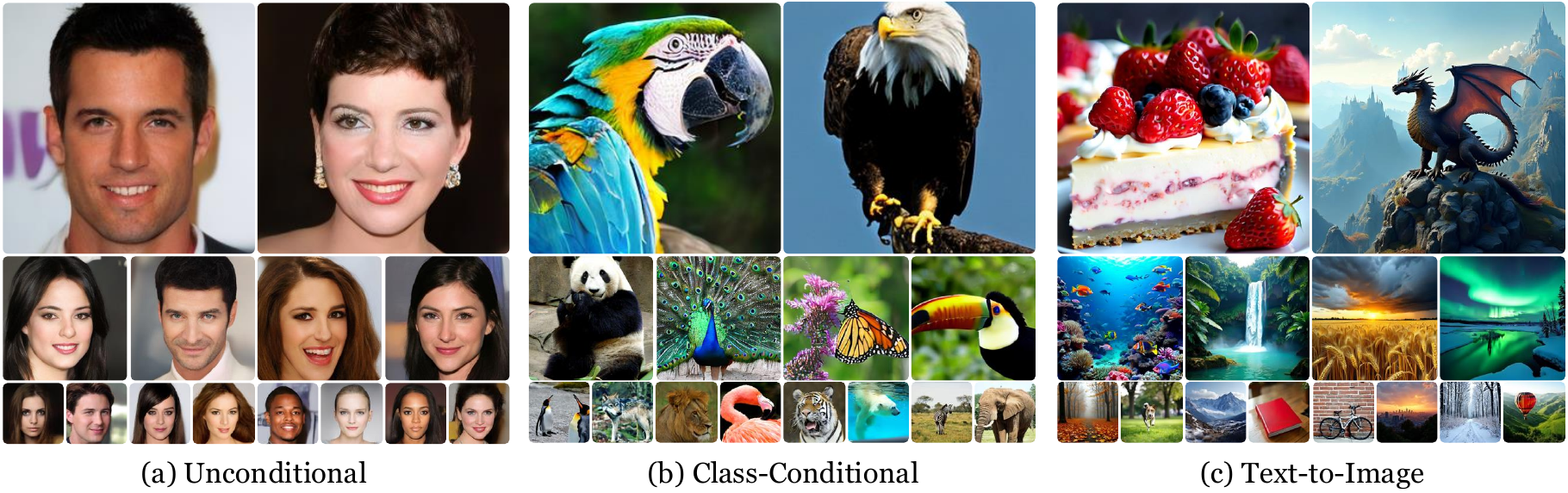}
    \vspace{-14pt}
    \caption{\textbf{Qualitative Examples.} \modelname produces high-quality and diverse samples across unconditional, class-conditional, and text-to-image generation settings.}
    \label{fig:qualitative}
\end{figure}

\begin{figure}[t]
\vspace{-5pt}
\begin{minipage}[t]{0.45\textwidth}
    \vspace{0pt}
    \centering
    \includegraphics[width=\linewidth]{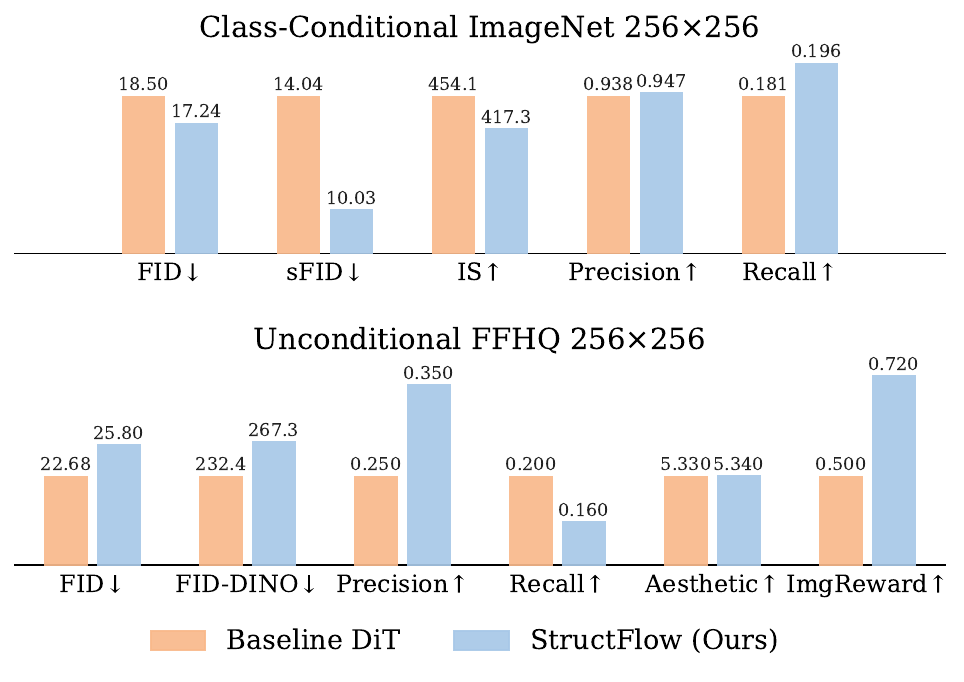}
    \vspace{-11pt}
    \caption{\textbf{Comparison with baseline.} Evaluated across a diverse set of metrics for robust assessment, \modelname performs on par with standard flow matching and outperforms it in some cases.}
    \label{fig:baseline_comparison}
\end{minipage}
\hfill
\begin{minipage}[t]{0.53\textwidth}
    \vspace{0pt}
    \centering
    % -------- First Table --------
    \footnotesize
    \setlength{\tabcolsep}{1pt}
    \captionof{table}{Ablation of training techniques
    % \mk{recall is missing bold value}
    }
    \label{tab:training_ablation}
    \vspace{-4pt}
    \begin{tabular}{lccccc}
    \toprule
    \multicolumn{6}{c}{\textbf{Class-Conditional ImageNet 256$\times$256}} \\
    \midrule
    Training Setting {\tiny(1M steps)} & FID$\downarrow$ & sFID$\downarrow$ & IS$\uparrow$ & Precision$\uparrow$ & Recall$\uparrow$ \\
    \midrule
    Baseline & 18.12 & 14.51 & 120.26 & 0.67 & \textbf{0.23} \\
    + Cosine Anchoring & 11.52 & 11.27 & 241.95 & 0.80 & 0.19 \\
    + Progressive Annealing & 10.97 & 10.43 & 330.00 & 0.85 & 0.18 \\
    + Mixed Coherence & \textbf{10.48} & \textbf{10.24} & \textbf{346.51} & \textbf{0.86} & 0.19 \\
    \bottomrule
    \end{tabular}

    \vspace{2pt}
    \includegraphics[width=0.90\linewidth]{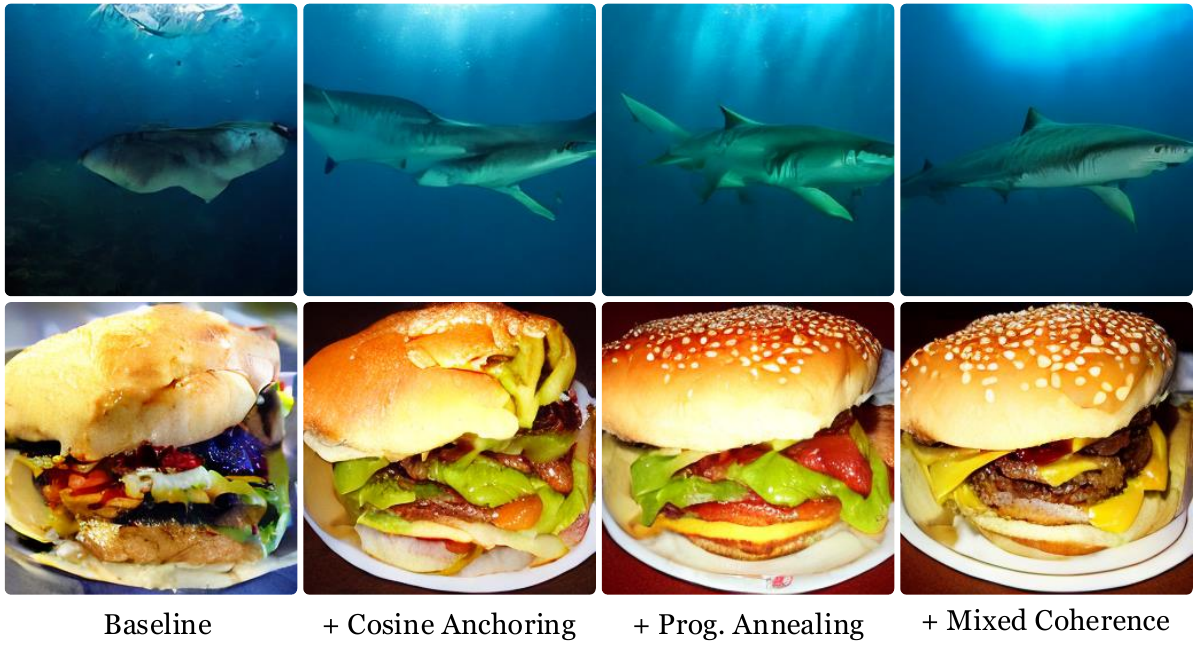}
    % -------- Second Table --------
    % \footnotesize
    % \setlength{\tabcolsep}{2.4pt}
    % \label{tab:training_ablation_ffhq}
    % \begin{tabular}{lcccc}
    % \toprule
    % \multicolumn{5}{c}{\textbf{Unconditional FFHQ 256$\times$256}} \\
    % \midrule
    % Training Setting & FID$\downarrow$ & Precision$\uparrow$ & Recall$\uparrow$ & Aesthetic$\uparrow$ \\
    % \midrule
    % Baseline & 0.00 & 0.00 & 0.00 & 0.00 \\
    % + Cosine Anchoring & 0.00 & 0.00 & 0.00 & 0.00 \\
    % + Progressive Annealing & 0.00 & 0.00 & 0.00 & 0.00 \\
    % + Mixed Coherence & 0.00 & 0.00 & 0.00 & 0.00 \\
    % \bottomrule
    % \end{tabular}
\end{minipage}
\end{figure}

We first evaluate how \modelname compares to standard flow matching (FM) when training DiT models. To ensure a fair comparison, we train models under identical settings (e.g., training iterations, batch size; see Appendix~\ref{sec:app:experiments:implementation_details} for full details) on both unconditional and class-conditional datasets.

Figure~\ref{fig:baseline_comparison} presents the quantitative comparison across a diverse set of metrics, including FID~\cite{heusel2017gans}, Inception Score (IS)~\cite{salimans2016improved}, DINO-based~\cite{oquab2023dinov2} FID, sFID~\cite{nash2021generating}, Precision/Recall~\cite{kynkaanniemi2019improved}, Aesthetic score~\cite{schuhmann2022laion}, and ImageReward~\cite{xu2023imagereward}. These metrics collectively capture both fidelity and diversity, enabling a comprehensive evaluation. As shown, \modelname performs on par with standard FM, and even surpasses it on several metrics. Specifically, on class-conditional ImageNet, \modelname matches or outperforms standard FM on FID, sFID, Precision, and Recall, while falling slightly behind on Inception Score. For unconditional FFHQ, \modelname similarly performs on par with standard FM on FID and Aesthetic score, while outperforming it on several metrics, including Precision and ImageReward. This demonstrates that \modelname preserves the strong generative quality of FM while enabling additional structured capabilities. Figure~\ref{fig:qualitative} further illustrates qualitative samples from \modelname-trained models across unconditional, class-conditional, and text-conditioned (see Sec.~\ref{sec:experiments:t2i}) settings. The generated images exhibit high visual quality and diversity, confirming that \modelname does not compromise sample realism while unlocking emerging capabilities. Refer to Appendix~\ref{app:sec:qualitative_and_quantitative} for more details.

We conduct an ablation study to analyze the contribution of each training strategy proposed in Sec.~\ref{sec:training_strategies}. Starting from the naive hierarchical noise sampling (Sec~\ref{sec:method:base_noise_sampling}), we incrementally introduce each component while keeping all other training settings fixed. All models are trained for 1M iterations to ensure a consistent comparison. As shown in Table~\ref{tab:training_ablation} with the qualitative samples, each added component improves training stability and overall sample quality, with the full model achieving the best performance. These results highlight the importance of our design choices and demonstrate that the proposed techniques lead to more reliable training and better generative performance.

\subsection{Applications}
\label{sec:experiments:applications}
In this section, we highlight the emerging capabilities and interesting behaviors that \modelname enables out of the box. These experiments are not intended to position \modelname as a specialized method
for image editing, layout control, or representation learning. Instead, they illustrate
properties that arise naturally from introducing spatial structure into the source distribution.
% \mk{we should clarify why there is no comparisons for these applications with other existing methods. this is to prevent reviewers from picking on us}

\begin{figure}[t]
\begin{minipage}[t]{0.49\textwidth}
    \centering
    \includegraphics[width=\linewidth]{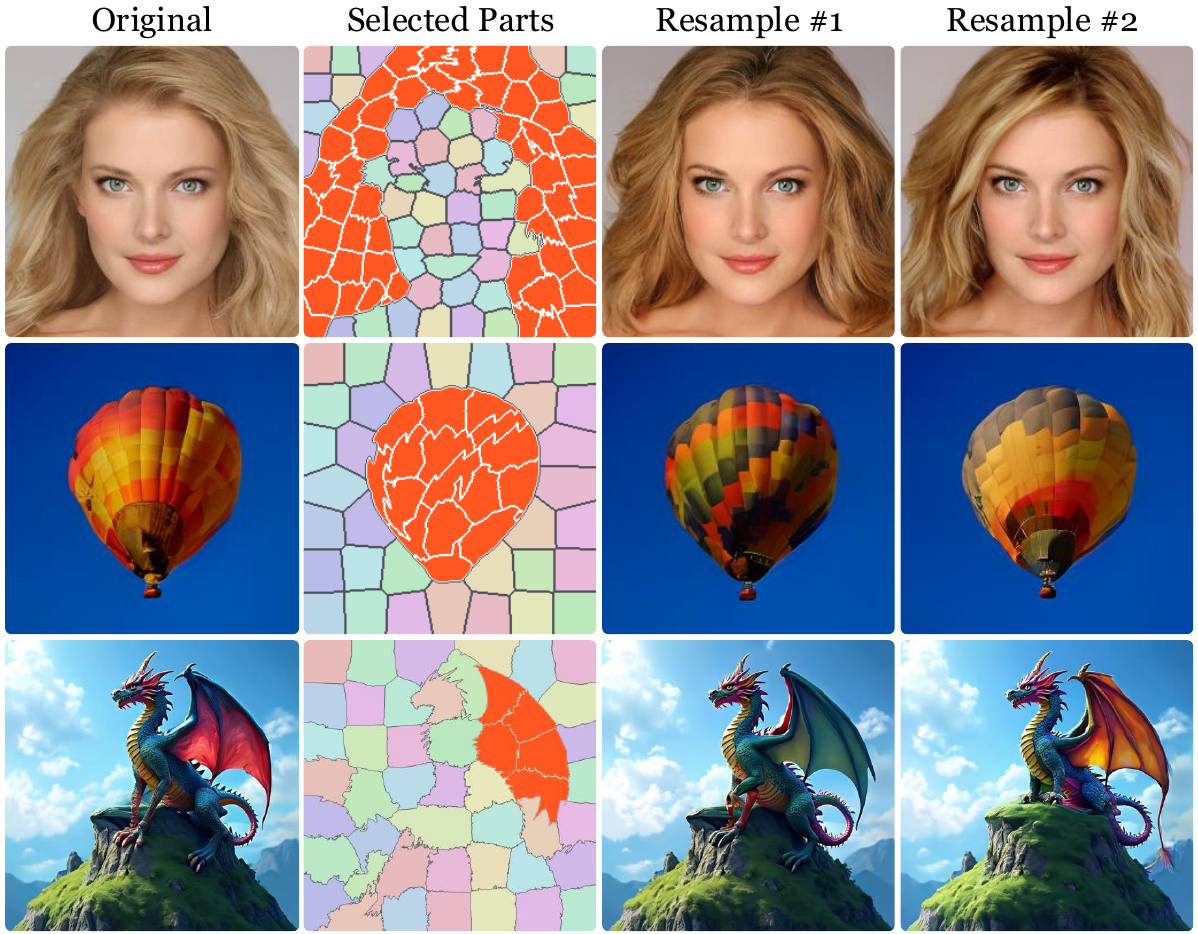}
    \vspace{-9pt}
    \caption{\textbf{Fine-Grained Editing.} \modelname allows fine-grained and local edits in the generations by resampling the noise in the selected local regions while keeping it fixed elsewhere.
    % \mk{you should add the diff between resampled and original stacked below the rgb images. original and selected parts on top of one another adjacent to resampled/diff}
    }
    \label{fig:resampling}
\end{minipage}
\hfill
\begin{minipage}[t]{0.49\textwidth}
    \centering
    \includegraphics[width=\linewidth]{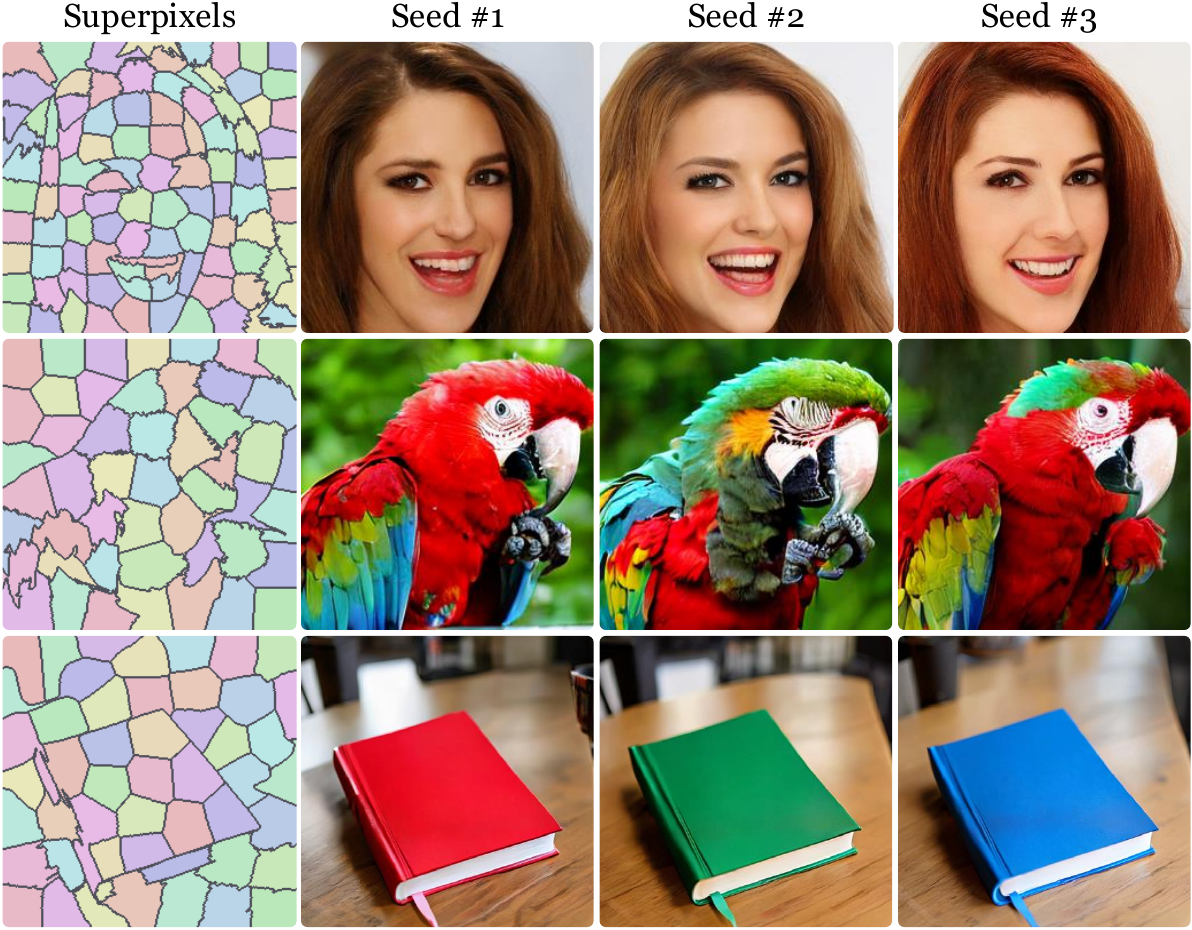}
    \vspace{-9pt}
    \caption{\textbf{Structure Preserving Generation.} \modelname allows generation of diverse samples using the same grounding superpixel masks while preserving the overall structure.
    % \mk{make sure the prompts are identical and only seed changes OR have two versions of this figure with one explicitly changes the seed while other only changes parts of the prompt like color of the book}
    }
    \label{fig:structure_preserving}
\end{minipage}
\end{figure}

\subsubsection{Fine-Grained Image Editing and Part Resampling}
\label{sec:part_resampling}

A key property of \modelname is that all pixels within the same segment $k$ share a common noise component, making different segments more clearly disentangled from one another. This structured decomposition enables targeted manipulation at the segment level. In particular, we can selectively resample the secondary noise for a subset of segments $\mathcal{S} = \{k_1, \dots, k_s\}$, which can be specified either by the user or automatically (e.g., via a segmentation model). By resampling only these segments while keeping the rest fixed, \modelname generates diverse variations within the selected regions while preserving the remaining content. Unlike SDEdit-style masked editing \cite{meng2021sdedit}, which preserves unmasked regions by explicitly replacing them at each denoising step, \modelname achieves localized resampling naturally through its structured source distribution without requiring explicit masking operations during denoising.
% \mk{emphasize how this is different from other methods which regenerate inside a mask}

Figure~\ref{fig:resampling} illustrates this behavior. The first column shows the initial generations, the second column highlights the selected regions, and the third and fourth columns present different resampled outputs for those regions. This enables precise, fine-grained local edits while preserving the integrity of other regions. For real images, this process can be applied by first inverting the image into the latent space using standard inversion techniques (e.g., DDIM~\cite{song2020denoising} or RF inversion~\cite{rout2025semantic}), and then performing segment-wise resampling in the same manner. Additional details are provided in Appendix~\ref{sec:app:applications:resampling}.

\subsubsection{Structure Preserving Generation}
\label{sec:structure_preserving_generation}

\modelname also enables explicit control over the spatial structure of generated images through its reliance on segmentation masks. In particular, the model naturally adheres to the provided segment boundaries, resulting in generations that respect the underlying structure.
This property allows users to design or modify segmentation masks to guide the global layout of the image. As a result, multiple samples generated from the same mask share a consistent structural composition, while still exhibiting diversity in appearance and fine details.
Figure~\ref{fig:structure_preserving} demonstrates this behavior. While all generated samples follow the same overall structure, they vary in texture, color, and local details, highlighting \modelname’s ability to preserve structure without sacrificing diversity. 
% Refer to Appendix~\ref{sec:app:applications:structure_preserving} for more examples.

\begin{figure}[t]
\begin{minipage}[t]{0.62\textwidth}
    \centering
    \includegraphics[width=\linewidth]{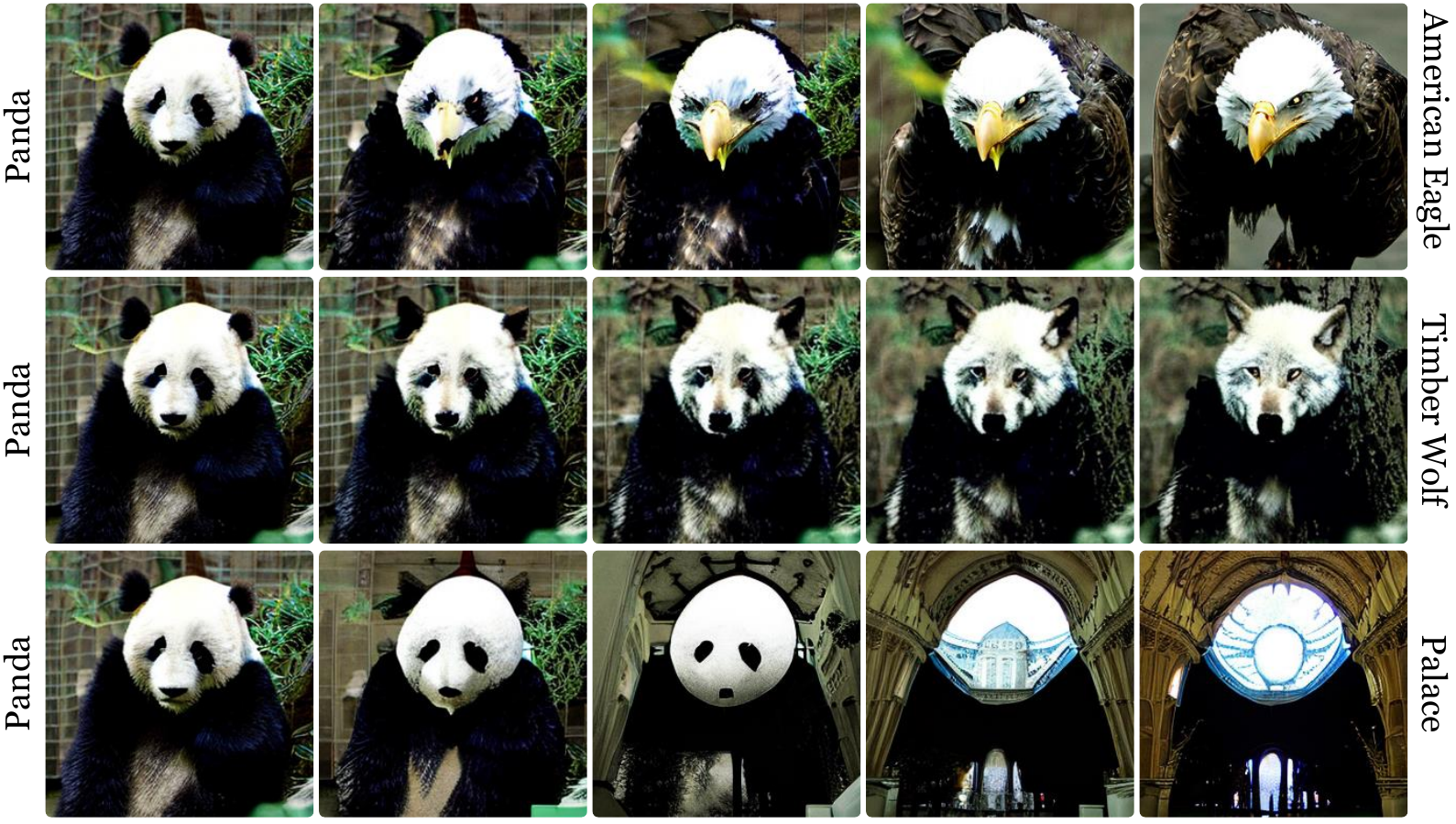}
    \vspace{-11pt}
    \caption{\textbf{Class Interpolation with Structure Preserving Generations.} \modelname's structure preserving capabilities allows interesting class interpolation during the early denoising steps of the generation.}
    \label{fig:class_interpolation}
\end{minipage}
\hfill
\begin{minipage}[t]{0.35\textwidth}
    \centering
    \includegraphics[width=\linewidth]{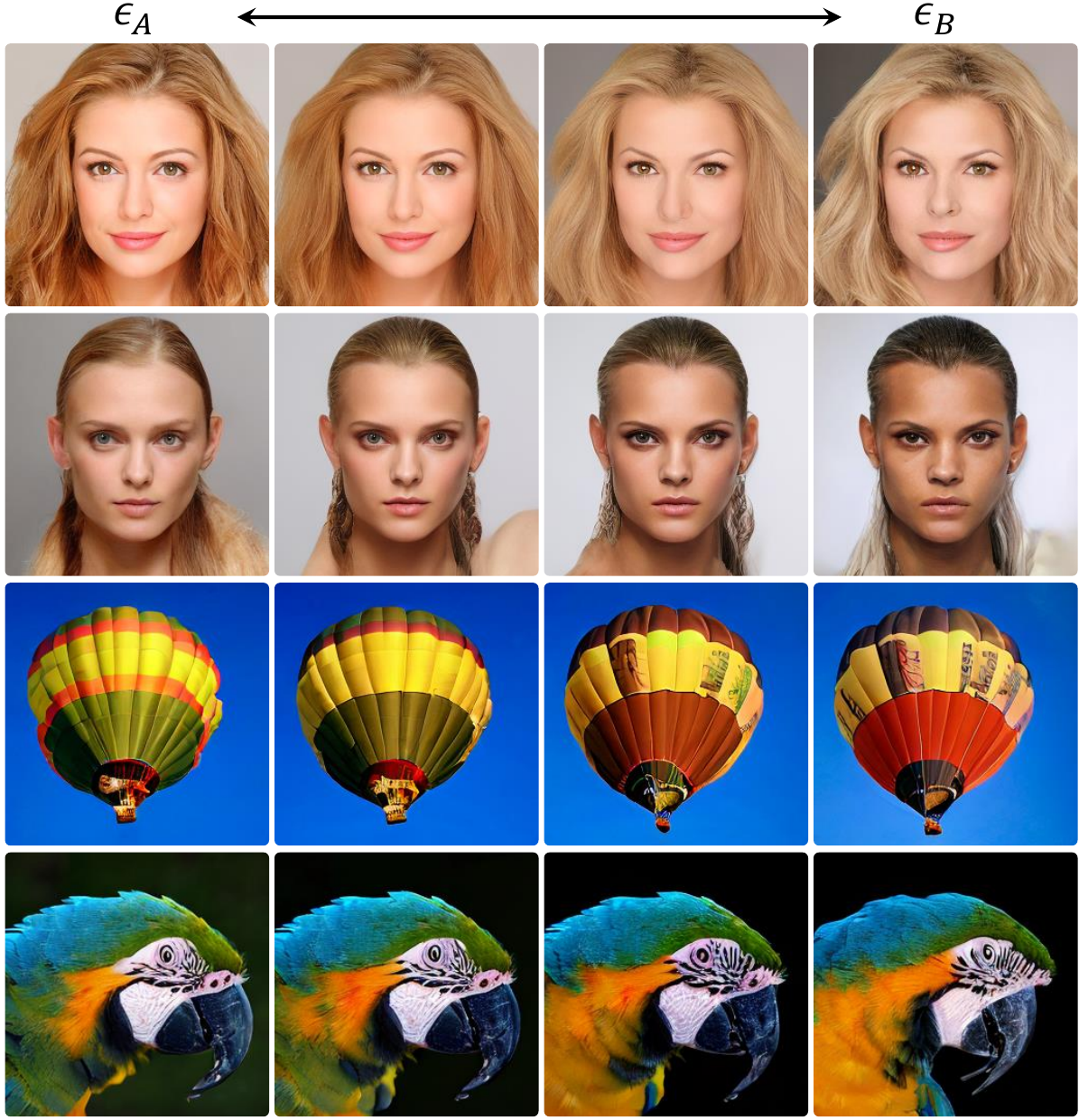}
    \vspace{-11pt}
    \caption{\textbf{Smooth Latent Interpolation.} Interpolating the initial generation noise ($e_{A,B}$) in \modelname yields smooth image transitions.
    % \mk{explain what epsilons mean here. also, can we have more drastic changes with more interpolation steps}
    }
    \label{fig:latent_interpolation}
\end{minipage}
\end{figure}

Figure~\ref{fig:class_interpolation} shows qualitative examples of \modelname trained on ImageNet, where the class label is swapped at different early timesteps during generation. The left column presents the original samples using the source class, while the subsequent columns show results when switching to the target class at progressively earlier stages. Beyond the smooth semantic transition, \modelname preserves the underlying spatial structure across classes, enabling consistent and meaningful class-level interpolations. Additional examples are provided in Appendix~\ref{sec:app:applications:structure_preserving}.

\subsubsection{Early Emergence of Informative Features}
\label{sec:informative_features}
\begin{figure}[t]
\begin{minipage}[t]{0.6\textwidth}
    \centering
    \includegraphics[width=\linewidth]{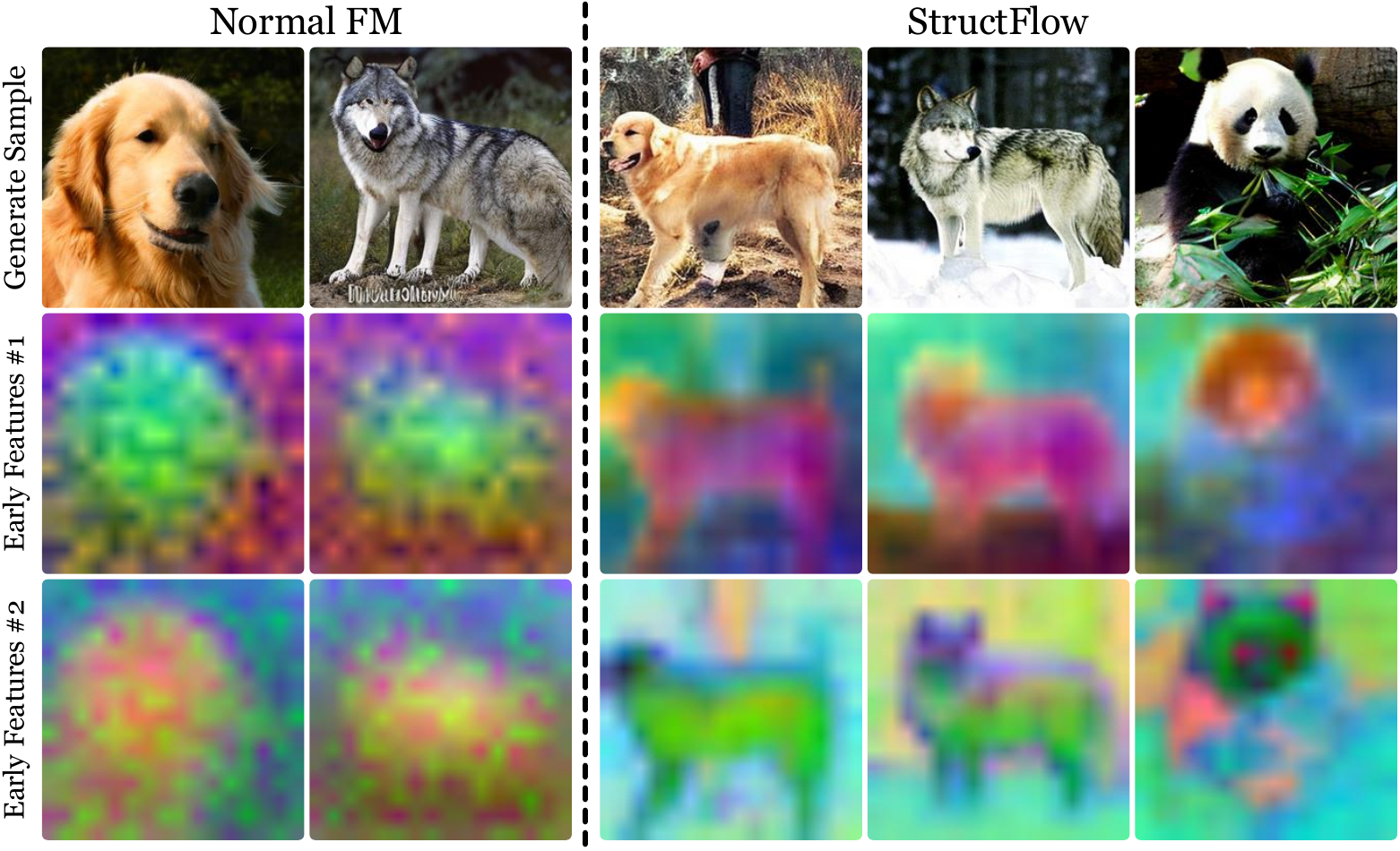}
    \vspace{-9pt}
    \caption{\textbf{Informative Early Features.} \modelname's internal features in early denoising inference steps show emergent informative features that can be used for downstream application such as segmentations.}
    \label{fig:pca_feature_interpretability}
\end{minipage}
\hfill
\begin{minipage}[t]{0.37\textwidth}
    \centering
    \includegraphics[width=\linewidth]{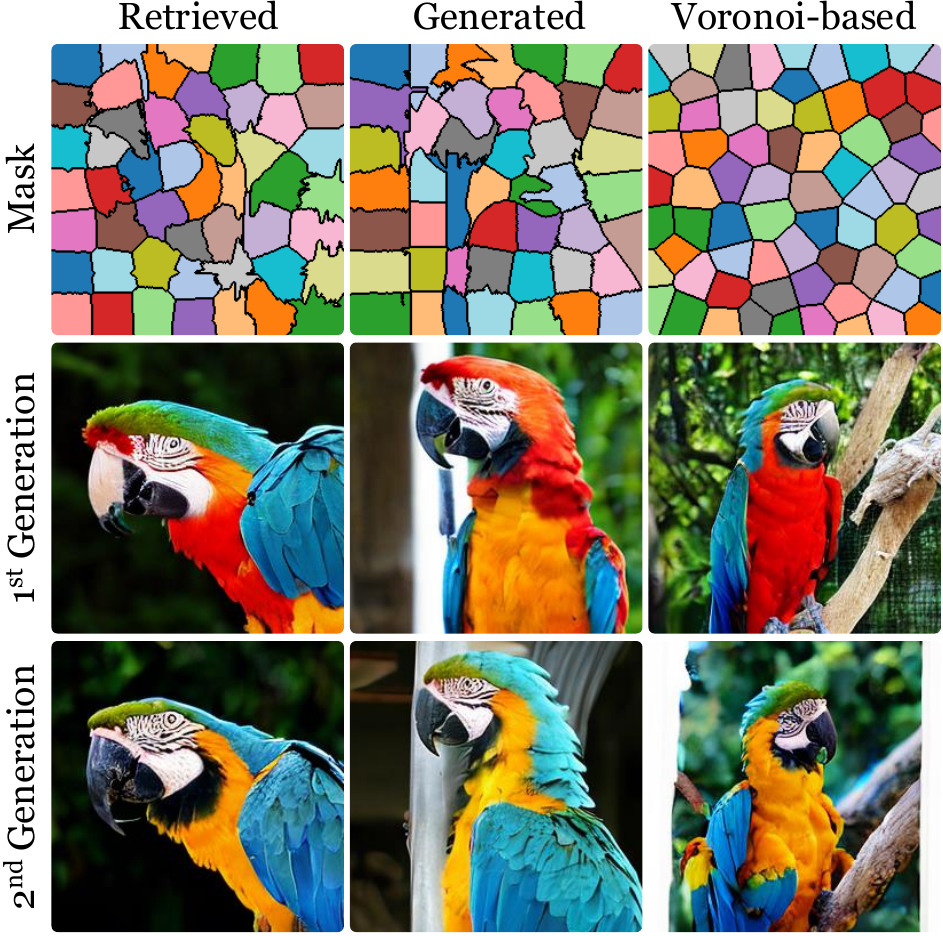}
    \vspace{-9pt}
    \caption{\textbf{Mask Generation.} We show that \modelname DiT models produce high-quality results across different mask generation methods.}
    \label{fig:mask_generation_ablation}
\end{minipage}
\end{figure}

We analyze the intermediate representations of DiT models during the denoising process by recording features at selected layers and timesteps across multiple generations. We then apply PCA to these features and visualize the top three components as RGB images. Interestingly, \modelname-trained models exhibit highly informative and spatially structured features even at very early timesteps. In particular, coarse object boundaries and layout emerge much earlier compared to models trained with standard flow matching. Figure~\ref{fig:pca_feature_interpretability} illustrates this behavior. Refer to Appendix~\ref{sec:app:applications:informative_features} for more details.

\subsubsection{Latent Interpolation}

We study the structure of the \modelname source distribution by sampling two noise instances and linearly interpolating between them, followed by generation from the interpolated latents.
As shown in Figure~\ref{fig:latent_interpolation}, this results in smooth and coherent transitions in the generated images. This behavior indicates that the latent space induced by \modelname is well-structured and semantically meaningful, enabling consistent interpolation between samples.

\subsection{Text-to-Image \modelname}
As discussed in 
\label{sec:experiments:t2i}Sec.~\ref{sec:method:progressive_annealing}, Progressive Coherence Annealing enables post-training of models originally trained with the standard flow matching objective. Leveraging this property, we post-train the SANA text-to-image model~\cite{xie2024sana} within the \modelname framework.
Qualitative results in Figure~\ref{fig:qualitative} (c) demonstrate high-fidelity image generation, while Figures~\ref{fig:structure_preserving} and~\ref{fig:resampling} highlight key capabilities such as accurate structure preservation and fine-grained editing. Refer to Appendix~\ref{sec:app:t2i} for more details.

\subsection{Mask Generation}
\label{sec:mask_generation}
\modelname relies on segmentation masks during training, readily obtained using superpixel methods such as SLIC~\cite{achanta2012slic}. At inference time, however, such masks are not directly available. Here, we outline several simple strategies to address this, with additional details provided in Appendix~\ref{sec:app:mask_generation}.

First, we consider a retrieval-based approach, where masks are sampled from a pool of precomputed masks. Given that \modelname can generate diverse outputs from a fixed mask, this already enables diverse image generation. Second, one can train a lightweight generative model (e.g., a small GAN or diffusion/flow model) to produce masks. These generated masks can then be used to construct the input noise. While effective, this approach introduces an additional model.
To avoid this, we also propose training-free alternatives. In particular, superpixel-like structures can be approximated by sampling seed points with a minimum-distance constraint (e.g., Poisson sampling), followed by assigning each pixel to its nearest seed to form a Voronoi tessellation. 
We further explore simple variants of such constructions and provide a comprehensive ablation study of training-free mask generation strategies in Appendix~\ref{sec:app:mask_generation}.
Figure~\ref{fig:mask_generation_ablation} visualizes these approaches, showing that all methods produce reasonable masks and lead to plausible generations, despite the model being trained only with SLIC masks.
Refer to Appendix~\ref{sec:app:mask_generation} for more details.
% \mk{do we want to show an example of actually running a semantic segmentation to create structure? this could work with FFHQ}

% \subsection{Ablations}
% effect of superpixel size and number, swapping superpixel with segmentation, ... \az{Maybe move to appendix}
\section{Conclusion}
We introduced StructFlow, a spatially grounded flow matching framework that replaces the standard i.i.d. Gaussian with a structured source distribution aligned with image locality. By correlating noise within local regions, StructFlow builds spatial coherence directly into the generative process, reducing the burden on the learned velocity field and enabling new forms of controllability. 
% To make this structured source practical, we proposed cosine-induced spatial anchoring, progressive coherence annealing, and mixed-coherence training, which together stabilize optimization and allow a flexible quality--editability trade-off.
Since naively training with such structured sources can be unstable, we proposed practical training strategies that stabilize optimization and preserve generation quality.
Across unconditional, class-conditional, and text-to-image settings, StructFlow demonstrate strong generation quality while enabling fine-grained local editing, structure-preserving generation, smooth interpolation, and more informative early representations. These results suggest that source distribution design is a powerful and underexplored axis for improving both the controllability and interpretability of generative models.

\bibliographystyle{abbrvnat}
\bibliography{ref}

\newpage
\appendix
\begin{figure}[t]
    \centering
    
    \begin{minipage}{\linewidth}
        \centering
        \includegraphics[width=\linewidth]{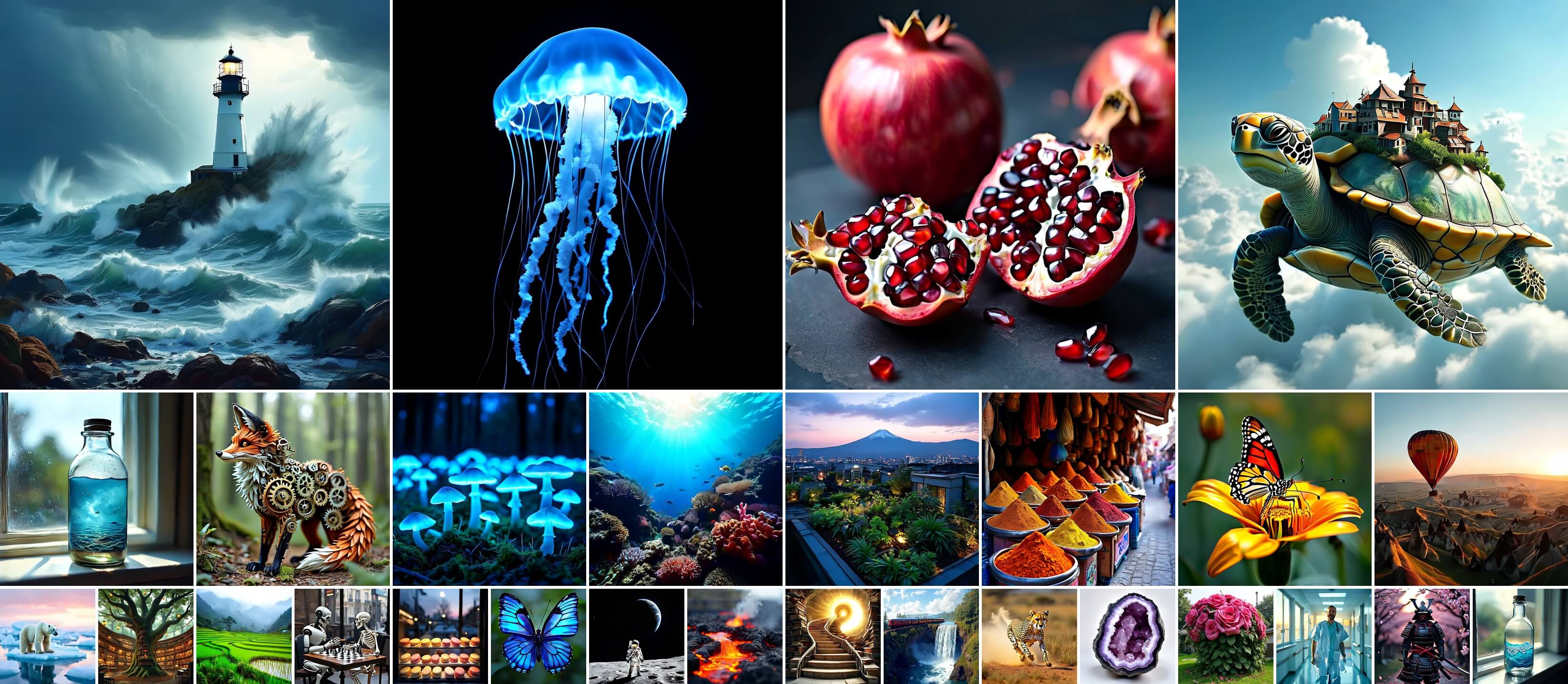}
        \captionof{figure}{\textbf{Text-to-Image Qualitative Examples}}
        \label{fig:appendix_t2i_qualitative}
    \end{minipage}
    
    \vspace{6pt}
    
    \begin{minipage}{\linewidth}
        \centering
        \includegraphics[width=\linewidth]{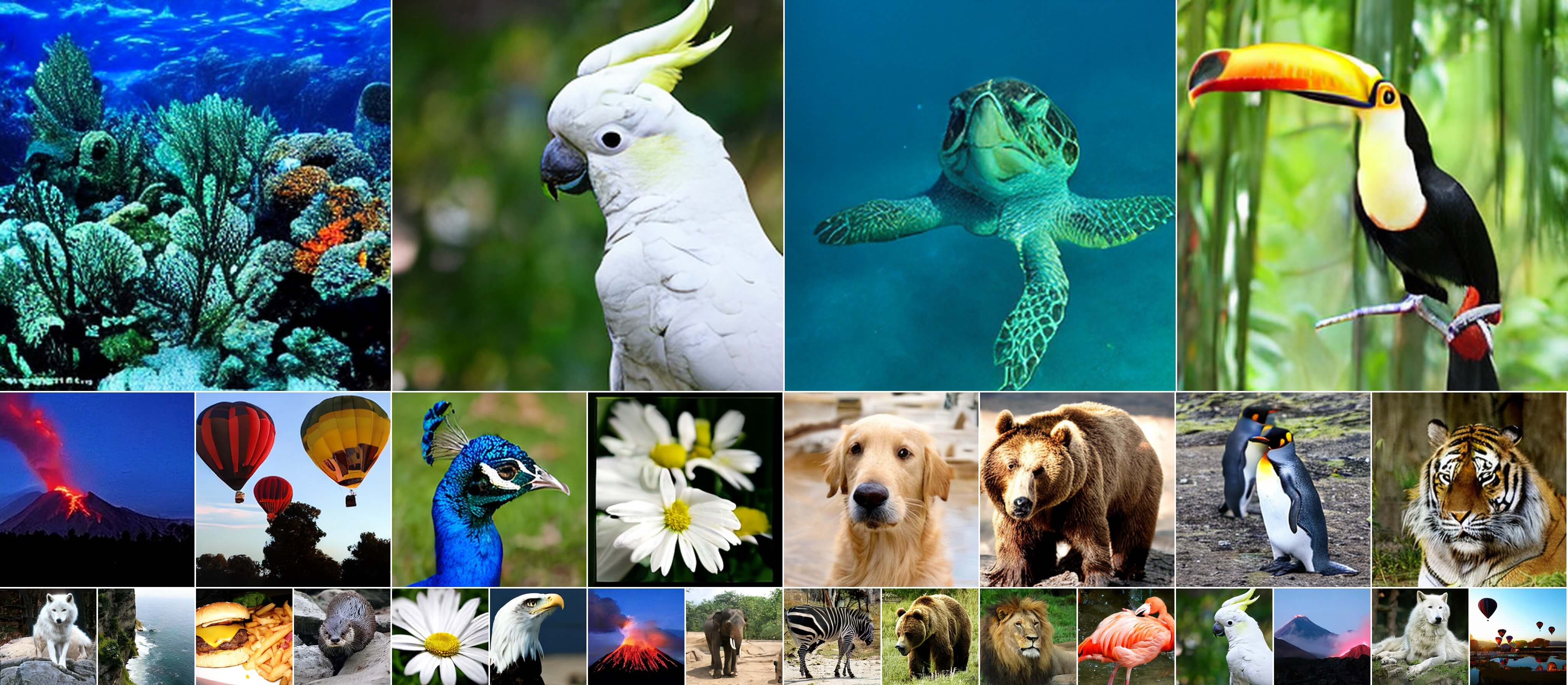}
        \captionof{figure}{\textbf{Class-Conditional ImageNet Qualitative Examples}}
        \label{fig:appendix_imagenet_qualitative}
    \end{minipage}
    
\end{figure}

\section{Related Works}
\subsection{Diffusion and Flow Models}
\label{app:related_works_diffusion_and_flow}

Diffusion models are a class of generative models grounded in stochastic differential equations (SDEs). Their core principle is to gradually corrupt data through a stochastic forward process that adds noise over time, ultimately transforming the data distribution into a simple Gaussian distribution. This forward process is formally defined as
$$dx = f(x, t) dt + g(t)dW_t,$$ 
where $f(x, t)$ denotes the drift term, $g(t)$ is the diffusion coefficient, and $dW_t$ represents the Wiener process, corresponding to infinitesimal Gaussian noise injected at time. The generative process corresponds to reversing this diffusion, aiming to recover the data distribution from noise. The reverse-time dynamics are given by
$$dx = [f(x, t) - g^2(t)\nabla_x\log p_t(x)]dt + g(t)dW_t,$$ 
where $\nabla_x \log p_t(x)$, known as the score function, captures the gradient of the log-density of the data distribution at time $t$. In practice, diffusion models learn this score function by training a neural network $s_\theta(x,t)$ via score matching:
$$\mathbb{E}_{t \sim U(0, T), x \sim p_t(x)}[\lambda(t)||\nabla_x\log p_t(x)-s_\theta(x,t)||^2],$$
where $\lambda(t)$ is a time-dependent weighting function. 

Closely related to diffusion models are flow matching methods, which provide a framework for training Continuous Normalizing Flows (CNFs). Unlike diffusion models, which rely on stochastic SDE trajectories, flow matching constructs deterministic mappings between distributions via ordinary differential equations (ODEs). Specifically, the data transformation is governed by a learned vector field:
$$\frac{dx}{dt}=v_\theta(x,t),$$
where $v_\theta(x,t)$ parameterizes the velocity field. The model is trained by minimizing the discrepancy between the learned vector field and a target vector field $v_t(x)$:
$$\mathbb{E}_{t \sim U(0, T), x \sim p_t(x)}[|v_\theta(x,t)-v_t(x)|^2],$$
where $p_t(x)$ denotes the intermediate distributions along the transport path.

In contrast to diffusion models, flow matching leverages deterministic ODE trajectories rather than stochastic SDE dynamics, which can lead to improved computational efficiency and more stable training. As a result, flow matching can be viewed as a principled and efficient alternative to diffusion-based generative modeling.

\section{Experiments}
\subsection{Implementation details}
\label{sec:app:experiments:implementation_details}

All models use DiT-XL/2~\cite{peebles2023scalable} as the backbone, operating on $32{\times}32$ latent representations obtained by encoding $256{\times}256$ images with a pretrained VAE (Stable Diffusion 
\texttt{sd-vae-ft-ema}~\cite{rombach2022high}, downsample factor 8). Training is performed on 8$\times$A100 GPUs using HuggingFace Accelerate~\cite{accelerate}, with bfloat16 mixed precision and no        
gradient checkpointing. We optimize with AdamW (lr~$= 10^{-4}$, weight decay $= 0$, $\varepsilon = 10^{-8}$) and a constant learning rate schedule. Gradient norms are clipped to 1.0 (ImageNet) and 2.0
(FFHQ). An exponential moving average (EMA) of model weights is maintained throughout training with decay $= 0.9999$; all evaluations use the EMA weights. Timesteps are sampled uniformly, and training    
follows the flow matching objective. Inference uses 100 Euler steps.

For StructFlow, segmentation masks are produced by SLIC superpixel oversegmentation~\cite{achanta2012slic} precomputed at full image resolution ($512{\times}512$) with 100 segments and compactness 30 for 
FFHQ, and 50 segments and compactness 50 for ImageNet. The structured noise is sampled at full resolution and downsampled to latent resolution ($32{\times}32$) with standard deviation rescaling to
preserve unit variance. The secondary noise strength $\lambda$ (which controls intra-segment coherence) is annealed during training via a piecewise linear schedule rather than held fixed, gradually       
tightening the segment constraint as training progresses. For FFHQ, $\lambda$ is annealed from 4.0 to 0.1 between steps 60k–100k and the model is trained for 130k steps total, using the 130k EMA
checkpoint for evaluation. For ImageNet, StructFlow training is conducted in sequential stages initialized from a pretrained flow matching checkpoint, with $\lambda$ annealed from 25.0 down to 4.0 and
subsequently to 0.5 across approximately 3M total iterations — matching the budget of the baseline.
The baseline DiT (FM-based) model is trained under an identical setup — same architecture, optimizer, batch size, training steps, and inference configuration — differing only in the noise sampling procedure, which uses standard isotropic Gaussian noise.

For the text-to-image experiments, we fine-tune SANA-1600M~\cite{xie2024sana} (1024px) on the FLUX-Reason-6M dataset~\cite{fang2025flux} using full fine-tuning (no LoRA). Training uses AdamW with lr~$= 2{\times}10^{-5}$,  
  $\beta = (0.9, 0.999)$, weight decay $= 10^{-4}$, and $\varepsilon = 10^{-8}$, with a constant learning rate schedule and no warmup. We use a global batch size of 32 ($4 \times 8$ GPUs), bfloat16 mixed   
  precision, no gradient checkpointing, and train for 100k steps. Caption dropout of 0.1 is applied to support classifier-free guidance at inference. Segmentation masks are produced on-the-fly via SLIC with
   50 segments and compactness 40 at full resolution ($1024{\times}1024$), then downsampled to latent resolution with std rescaling. As with ImageNet, the secondary noise strength $\lambda$ is annealed from
   22.0 to 0.5 between steps 5k–65k to progressively tighten the segment constraint during training.

\subsection{Qualitative and Quantitative evaluation}
\label{app:sec:qualitative_and_quantitative}

In this section, we provide additional qualitative examples from a DiT trained with StructFlow and a SANA model post-trained using our method. Figures~\ref{fig:appendix_t2i_qualitative} and~\ref{fig:appendix_imagenet_qualitative} illustrate results from these two models. As shown, both produce high-quality samples while exhibiting a range of emerging capabilities.

\section{Applications}
\label{sec:app:applications}

\subsection{Fine-Grained Image Editing and Part Resampling}
\label{sec:app:applications:resampling}

\begin{figure}[t]
    \centering
    \begin{minipage}{\linewidth}
        \centering
        \includegraphics[width=\linewidth]{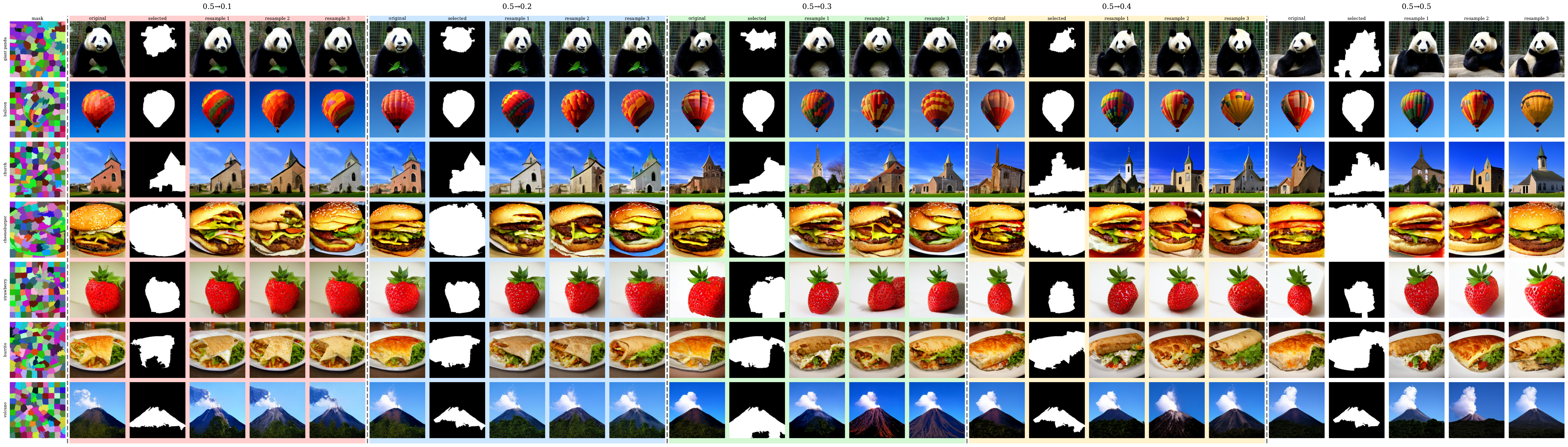}
        \captionof{figure}{\textbf{ImageNet Fine-Grained Local Resampling Qualitative Comparison.} Comparison across models trained with different $\lambda_\varsigma$. Models trained with smaller $\lambda_\varsigma$ preserve unintended regions while producing diverse variations within the selected parts.}
        \label{fig:appendix_qualitative_resampling_comparison_imagenet}
    \end{minipage}
    
    \vspace{6pt}
    
    \begin{minipage}{\linewidth}
        \centering
        \includegraphics[width=\linewidth]{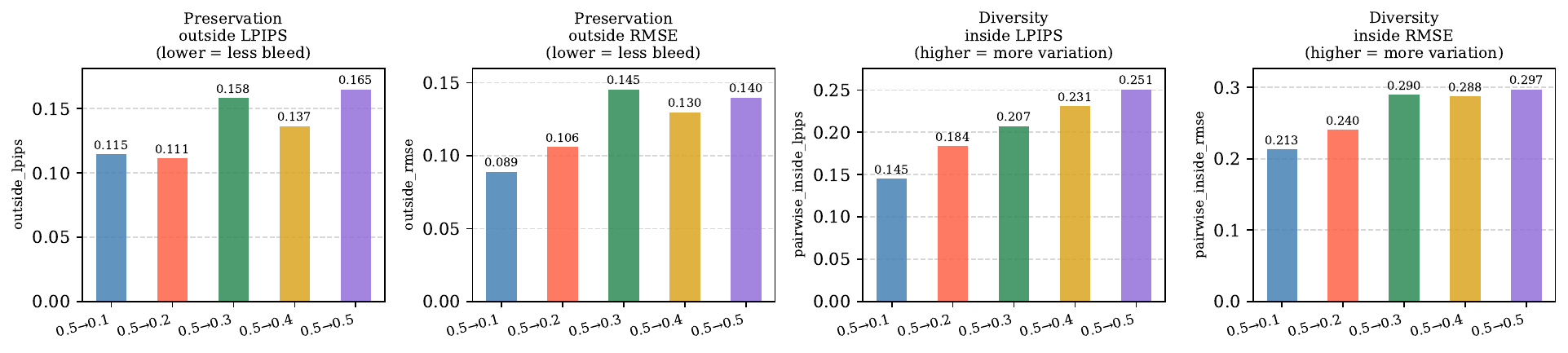}
        \captionof{figure}{\textbf{ImageNet Fine-Grained Local Resampling Quantitative Comparison.} Quantitative results corroborate the qualitative findings: models trained with smaller $\lambda_\varsigma$ better preserve unaffected regions while maintaining diversity within the selected parts.}
        \label{fig:appendix_quantitative_resampling_comparison_imagenet}
    \end{minipage}
    
\end{figure}

  In this section, we thoroughly evaluate the ability of fine-grained local resampling in \modelname. This is the ability to locally re-synthesize a specific semantic region of an image while leaving the rest unchanged. Concretely, given a generated image and its associated oversegmentation mask, we identify a semantically meaningful subset of 
  segments, regenerate only those segments under fresh noise, and measure both how well the unselected region is preserved and how diverse the regenerated variants are.                 
  \paragraph{Procedure.}          
  For each evaluation sample, a base image is first synthesized together with a SLIC oversegmentation mask~\cite{achanta2012slic} computed at the original image resolution. A region of interest is then identified automatically: we run CLIPSeg~\cite{luddecke2022image}
   on the base image using a text query, obtain a per-pixel relevance heatmap, average the heatmap values within each segment, and threshold the per-segment scores via Otsu's method~\cite{otsu1979threshold} (with a minimum score floor of 0.3) to obtain the set of selected segment IDs. For ImageNet models the query is the ground-truth class 
  name of the conditioning label (e.g., \emph{``golden retriever''}), so the selected region reflects the primary object of interest. For FFHQ faces the query is fixed to \emph{``hair''}. Given the selected region mask, we produce $K=5$ resample variants: each variant uses the same primary noise as the base image for the unselected pixels,
  but draws fresh secondary noise only inside the selected region, so that all stochasticity outside it is held fixed. This directly mirrors the structured noise decomposition of our \modelname training and provides a clean isolation of the regenerated zone.        
  \paragraph{Metrics.} 
  We measure two complementary properties across the $K$ resamples:                                                                                
We propose two complementary metrics to evaluate this behavior:
\textbf{Preservation} (lower is better) measures how much the \emph{unselected} region changes between the base image and each resample. We report the mean pixel-level RMSE (\textit{Outside RMSE}) and perceptual LPIPS distance (\textit{Outside LPIPS})~\cite{zhang2018unreasonable}, computed over unselected pixels only.
\textbf{Diversity} (higher is better) measures how much the \emph{selected} region varies \emph{across} resamples. We compute all $\binom{K}{2}$ pairwise comparisons within the selected region and report the mean pairwise RMSE (\textit{Inside RMSE}) and LPIPS (\textit{Inside LPIPS}).
A well-behaved model should achieve both low preservation error (minimal disturbance to surrounding regions) and high diversity (meaningful variation within the edited region).
  
\begin{figure}[t]
    \centering
    \begin{minipage}{\linewidth}
        \centering
        \includegraphics[width=\linewidth]{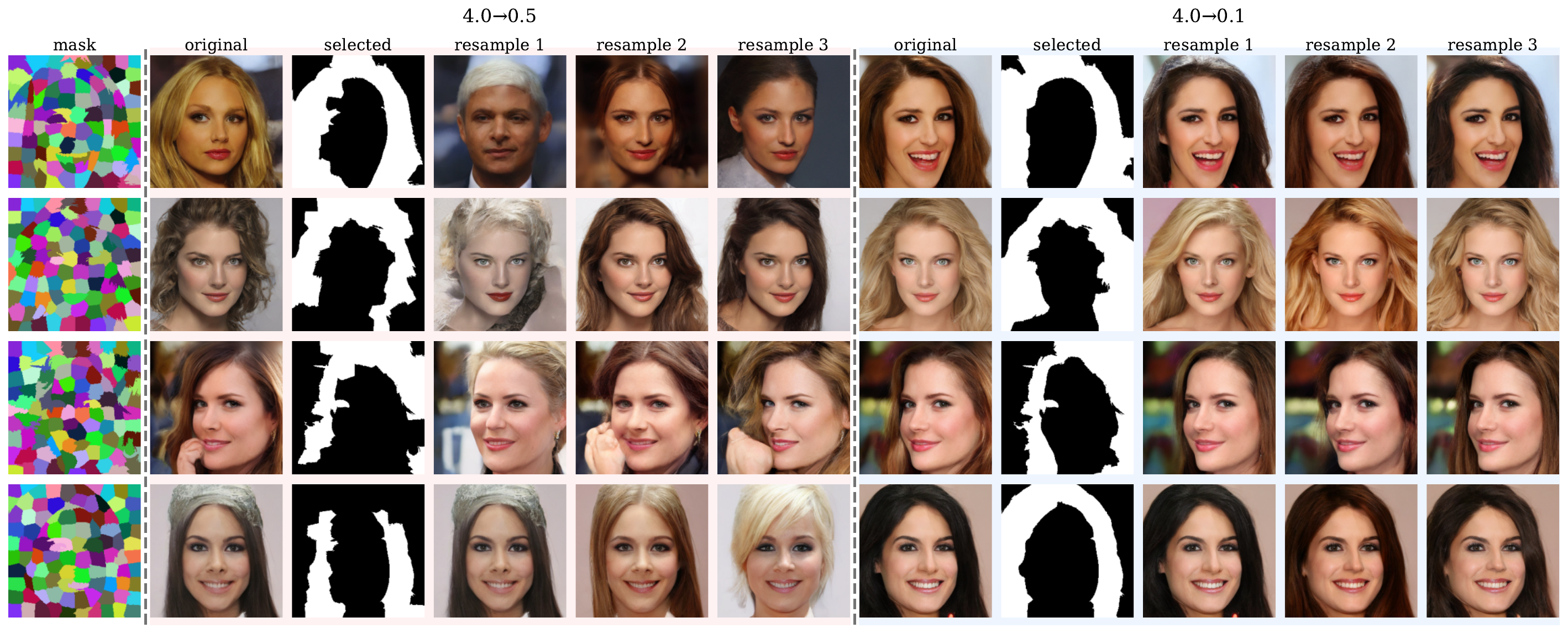}
        \captionof{figure}{\textbf{FFHQ Fine-Grained Local Resampling Qualitative Comparison.} Comparison across models trained with different $\lambda_\varsigma$. Models trained with smaller $\lambda_\varsigma$ preserve unintended regions while producing diverse variations within the selected parts.}
        \label{fig:appendix_qualitative_resampling_comparison_ffhq}
    \end{minipage}
    
    \vspace{6pt}
    
    \begin{minipage}{\linewidth}
        \centering
        \includegraphics[width=\linewidth]{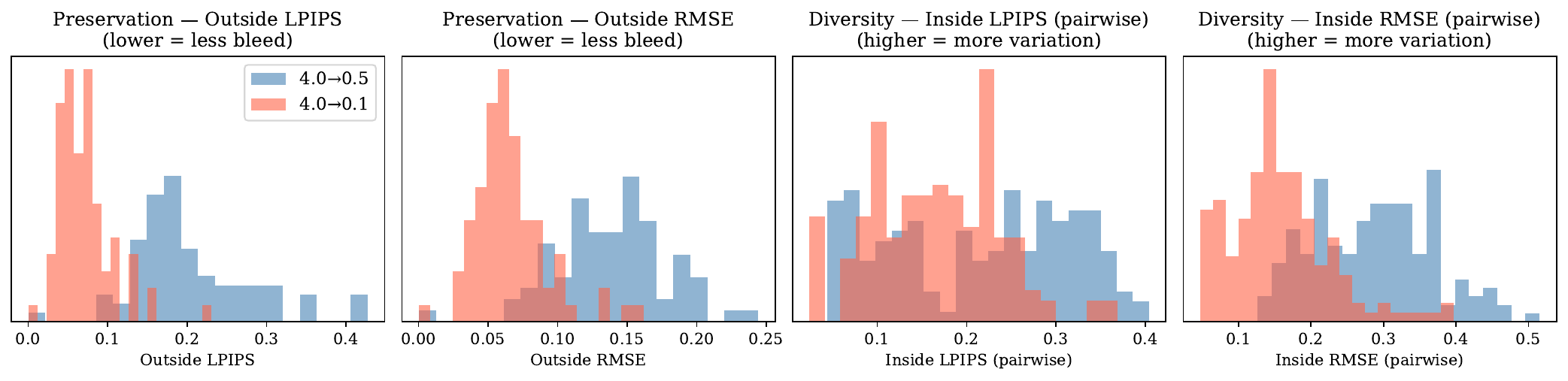}
        \captionof{figure}{\textbf{FFHQ Fine-Grained Local Resampling Quantitative Comparison.} Quantitative results corroborate the qualitative findings: models trained with smaller $\lambda_\varsigma$ better preserve unaffected regions while maintaining diversity within the selected parts.}
        \label{fig:appendix_quantitative_resampling_comparison_ffhq}
    \end{minipage}
    
\end{figure}

  \paragraph{ImageNet Results.}   
  We evaluate five model variants, each fine-tuned from a shared checkpoint that was originally trained with $\lambda_\varsigma = 0.5$, continuing with a cosine schedule that decays $\lambda_\varsigma$ to a final value of $\{0.1, 0.2, 0.3, 0.4, 0.5\}$ respectively (no mixed-coherence is used in these experiments). Evaluation is performed on 16 class-conditional validation samples with     
  classifier-free guidance scale 4.0; the CLIPSeg query matches the class label of each sample. As shown in Figures~\ref{fig:appendix_qualitative_resampling_comparison_imagenet} and~\ref{fig:appendix_quantitative_resampling_comparison_imagenet}, models trained to smaller final $\lambda_\varsigma$ exhibit substantially better preservation of
   the unselected region, with only a modest reduction in inside diversity. This confirms the theoretical expectation: tighter coupling between pixels of the same segment (smaller secondary noise) produces cleaner segment boundaries and more faithful outside preservation during resampling, while still leaving enough degrees of freedom      
  \paragraph{FFHQ. Results}         
  We compare two unconditional face generation models that differ in their secondary noise schedule: one trained with $\lambda_\varsigma$ annealed from 4.0 to 0.5, and one with a more aggressive schedule from 4.0 to 0.1. Evaluation uses 16 real images with SLIC masks, and the query \emph{``hair''} is used for
  automatic region selection. Results in Figures~\ref{fig:appendix_qualitative_resampling_comparison_ffhq} and~\ref{fig:appendix_quantitative_resampling_comparison_ffhq} show the same trend as ImageNet: the model with the smaller final $\lambda_\varsigma$ (4.0→0.1) substantially outperforms on preservation, 
  indicating that the hair region can be re-synthesized with minimal leakage into the face and background, while the diversity of the generated hair variants remains adequate for practical use.

\subsection{Structure Preserving Generation}
\label{sec:app:applications:structure_preserving}

\begin{figure}[t]
\begin{minipage}[t]{0.56\textwidth}
    \centering
    \includegraphics[width=\linewidth]{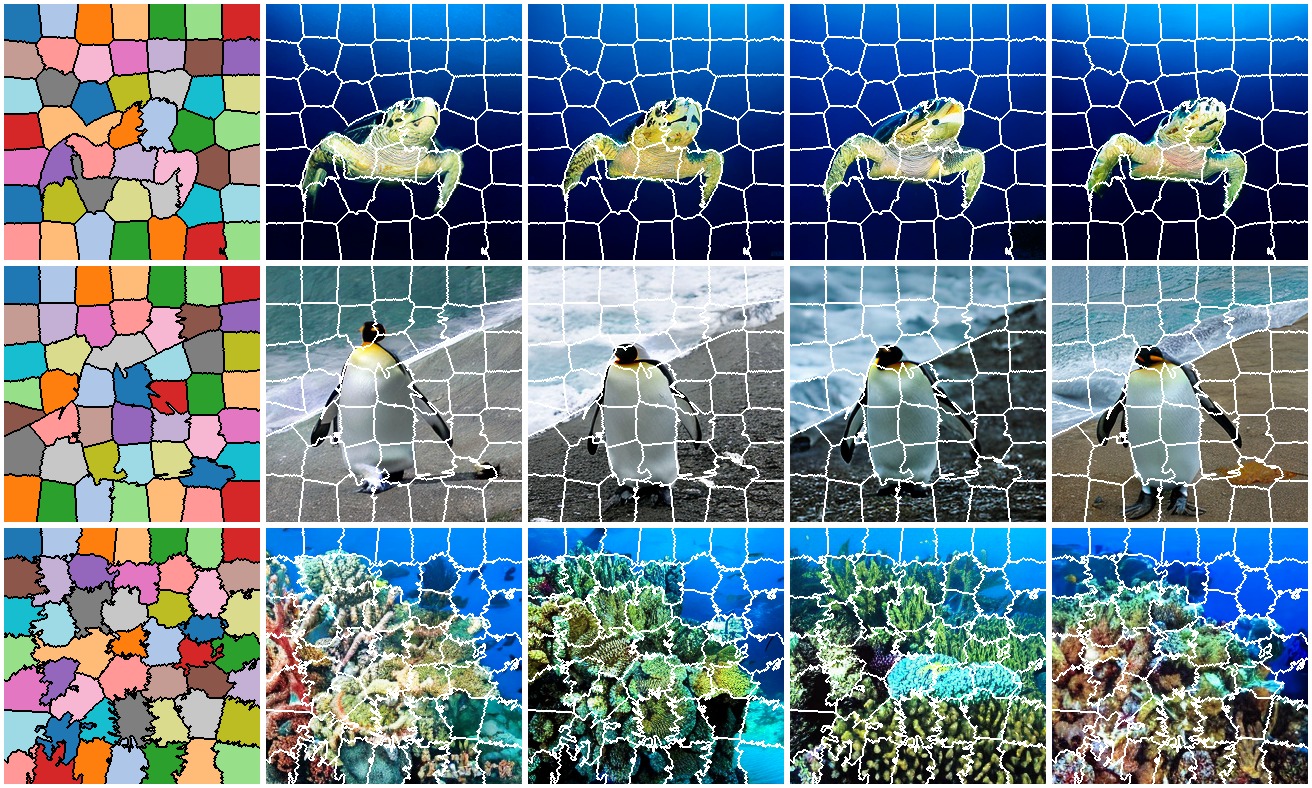}
    \vspace{-9pt}
    \caption{\textbf{Structure Preserving Generation.} \modelname allows generation of diverse samples using the same grounding superpixel masks while preserving the overall structure.}
    \label{fig:appendix_structure_preserving}
\end{minipage}
\hfill
\begin{minipage}[t]{0.42\textwidth}
    \centering
    \includegraphics[width=\linewidth]{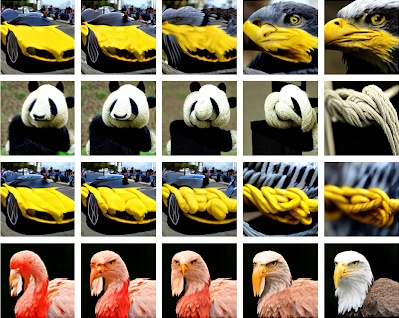}
    \vspace{-9pt}
    \caption{\modelname's structure preserving capabilities allows interesting class interpolation during the early denoising steps of the generation.}
    \label{fig:appendix_class_interpolation}
\end{minipage}
\end{figure}

In this section, we provide additional qualitative examples showing that \modelname naturally adheres to the provided segment boundaries, producing generations that preserve the underlying spatial structure. This structural consistency also enables class interpolation, where changing the class condition during early denoising alters the semantic content while largely maintaining the same layout. Figures~\ref{fig:appendix_structure_preserving} and~\ref{fig:appendix_class_interpolation} show additional results.

\begin{figure}[t]
    \centering
    \includegraphics[width=0.90\linewidth]{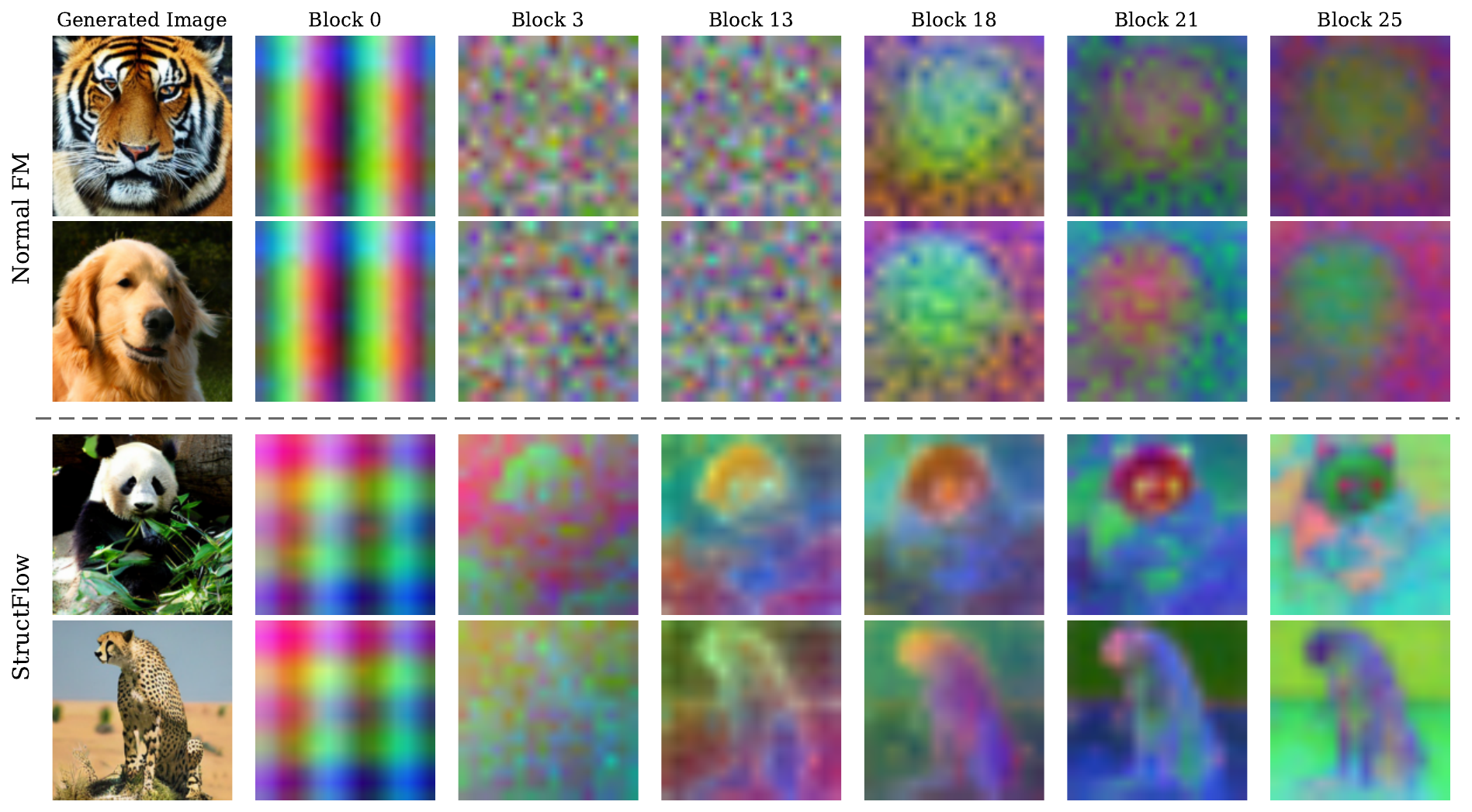}
    \caption{\textbf{Informative Early Features.} \modelname{} exhibits spatially coherent features from the earliest denoising steps, while Normal FM remains largely unstructured.}
    \label{fig:app:pca_features_appendix}
\end{figure}

\subsection{Early Emergence of Informative Features}
\label{sec:app:applications:informative_features}

We compare the intermediate representations of two DiT-XL/2 models trained on ImageNet
256$\times$256 and sampled with 100 Euler flow-matching steps using classifier-free guidance
(scale $=4.0$): (i) a standard flow-matching baseline, referred to as Normal FM, and
(ii) our \modelname{} model, whose source noise is constructed using SLIC oversegmentation
masks with 50 segments.
For each model, we generate $N=60$ images from 12 visually related animal classes, with
5 samples per class. These classes are chosen because they share common semantic parts, such
as eyes, ears, muzzles, and fur patterns, making cross-sample feature alignment easier to
inspect in the PCA maps. We extract features from all transformer blocks at five denoising
fractions: 10\%, 30\%, 50\%, 70\%, and 90\% of the sampling trajectory.
For each \emph{(block, timestep)} pair, we pool the spatial tokens from all 60 generated
images and fit a 3-component PCA. We then project each token onto the three principal
components, reshape the projected features back to the spatial grid, and upsample them to
the image resolution. The three PCA components are visualized as RGB channels and normalized
using the global minimum and maximum across all samples for that \emph{(block, timestep)}
pair. Therefore, within each pair, the same color corresponds to the same feature direction
across all samples, making cross-image consistency directly visible. Since PCA is fit
independently for each \emph{(block, timestep)} pair, colors should not be compared across
different rows or columns of the figure.

Figure~\ref{fig:app:pca_features_appendix} shows PCA maps from an early denoising stage,
at 10\% of the trajectory, i.e., after only 10 of the 100 sampling steps. We visualize six
representative transformer blocks. The first two rows correspond to Normal FM generations,
while the bottom two rows correspond to \modelname{} generations. The first column shows the
final generated image for reference.
Although the latent state is still highly noisy at this stage, the PCA maps of \modelname{}
already show clear spatial organization across multiple blocks, including coarse object
boundaries, foreground--background separation, and rough object layout. In contrast, the
Normal FM maps at the same timestep remain largely unstructured, with colors varying
irregularly across patches rather than forming coherent regions. For Normal FM, comparable
spatial organization appears only later in the denoising process, around 50\%--90\% of the
trajectory. This suggests that the structured source used by \modelname{} encourages the
model to organize its internal features around spatially meaningful regions from the earliest
sampling steps.

\section{Text-to-Image StructFlow}
\label{sec:app:t2i}

Details of the StructFlow text-to-image training setup are provided in Sec.~\ref{sec:app:experiments:implementation_details}. Additional qualitative examples are shown in Fig.~\ref{fig:appendix_t2i_qualitative}.

\section{Mask Generation Strategies}
\label{sec:app:mask_generation}

\begin{figure}[t]
    \centering
    \includegraphics[width=0.85\linewidth]{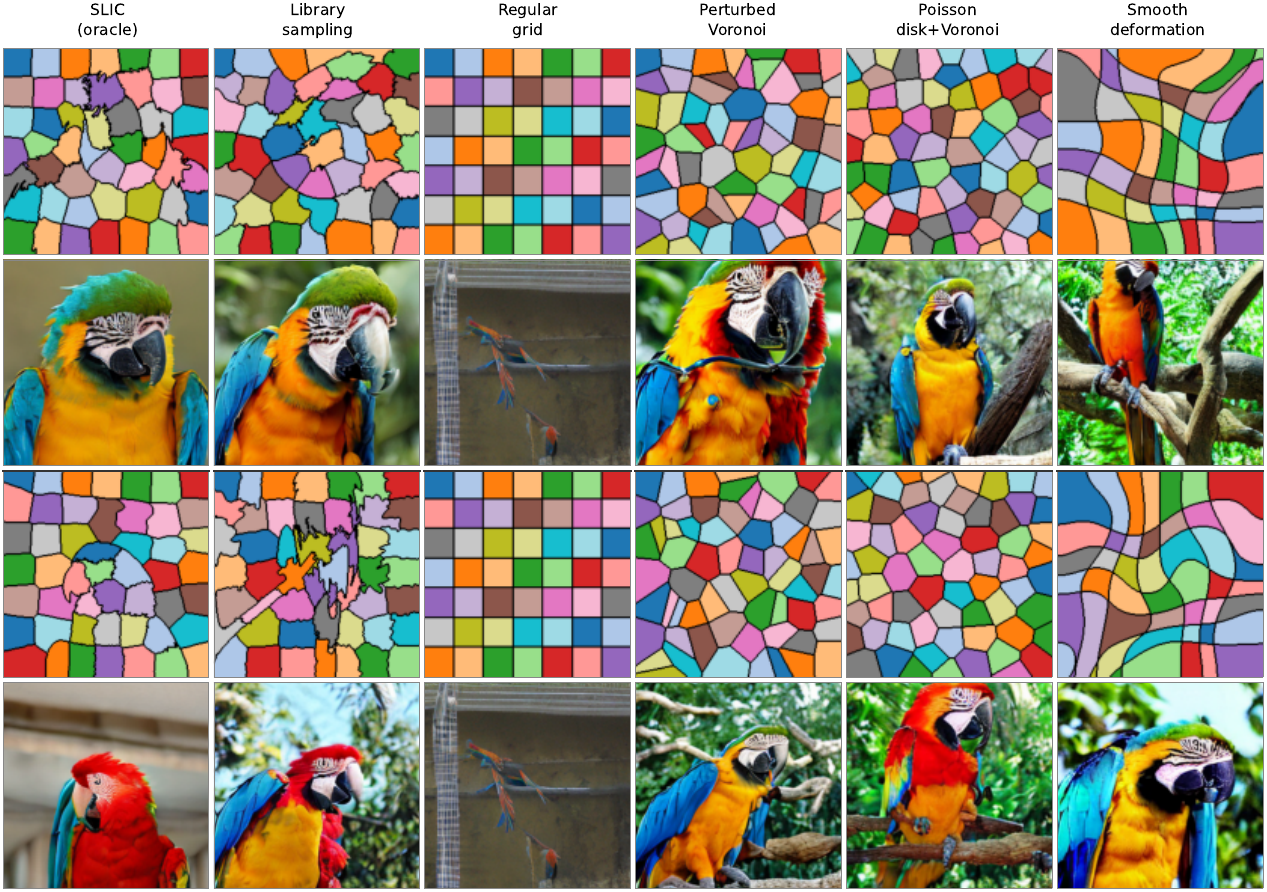}
    \caption{Comparison of training-free mask generation strategies}
    \label{fig:appendix_training_free_mask_ablation}
\end{figure}

As discussed in Section~\ref{sec:mask_generation}, we explore several strategies for
generating segmentation masks at inference time.
Here, we provide a more detailed discussion of these approaches.

\paragraph{Lightweight generative model.}
A straightforward option is to train a small generative model to produce plausible
superpixel layouts on demand.
We implement this as a compact U-Net flow-matching model (14.1M parameters) conditioned
on the ImageNet class label, trained on 1.28M precomputed SLIC masks in approximately
3 hours on two A100 GPUs.
The model closely matches real SLIC statistics in segment count, mean segment size, and
spatial autocorrelation.
Crucially, mask generation costs only 7--35\,ms per sample (batch sizes 32--128,
20 inference steps)---roughly two orders of magnitude faster than the main DiT image
generation pipeline---making it a negligible overhead in practice.

\paragraph{Training-free approaches.}
We explore five training-free alternatives requiring no additional model.
The simplest is \emph{library sampling}: at inference time, a mask is drawn uniformly at
random from the pool of precomputed SLIC masks.
Since SLIC superpixels with high compactness are governed primarily by spatial regularity
rather than image content, a mask from an unrelated image still provides a plausible
spatial layout.
This approach is zero-cost and produces in-distribution masks by construction.

The remaining strategies construct masks procedurally.
A \emph{regular grid} partitions the image into uniform rectangular cells.
\emph{Perturbed Voronoi} tessellation places seeds on a regular grid and applies isotropic
Gaussian noise before computing the nearest-neighbour Voronoi diagram; the perturbation
magnitude controls the degree of organic irregularity.
\emph{Poisson-disk Voronoi} removes the grid prior entirely: seeds are placed using
Bridson's algorithm~\cite{bridson2007fast}, which enforces a minimum inter-seed distance
and produces blue-noise point distributions with no underlying grid memory, yielding
more organic, irregular cells while still guaranteeing uniform spatial coverage.
Finally, \emph{smooth deformation} starts from a regular $n \times n$ grid and applies
a spatially smooth random displacement field---obtained by blurring independent Gaussian
noise with a large kernel---to warp the grid boundaries.

Figure~\ref{fig:appendix_training_free_mask_ablation} compares example masks and generated images for
all six strategies.
As the figure illustrates, the regular grid and smooth deformation approaches produce masks
that are noticeably out of distribution relative to the SLIC superpixels the model was
trained on: their large, rectangular or broadly warped regions differ substantially in
shape, size distribution, and boundary character from organic SLIC segments, and this
mismatch leads to visibly degraded generation quality.
By contrast, the Voronoi-based approaches---both perturbed Voronoi and Poisson-disk
Voronoi---produce masks whose local statistics are much closer to SLIC, and the resulting
generations are correspondingly higher quality and more coherent.
Library sampling inherits the exact SLIC distribution by construction and therefore
performs comparably to the oracle.
These results confirm that matching the mask distribution seen during training is the key
factor for high-quality generation, and that Voronoi-based procedural masks provide a
practical and effective training-free alternative to precomputed SLIC masks.

\section{Ablations}

\begin{figure}[t]
    \centering
    \includegraphics[width=0.95\linewidth]{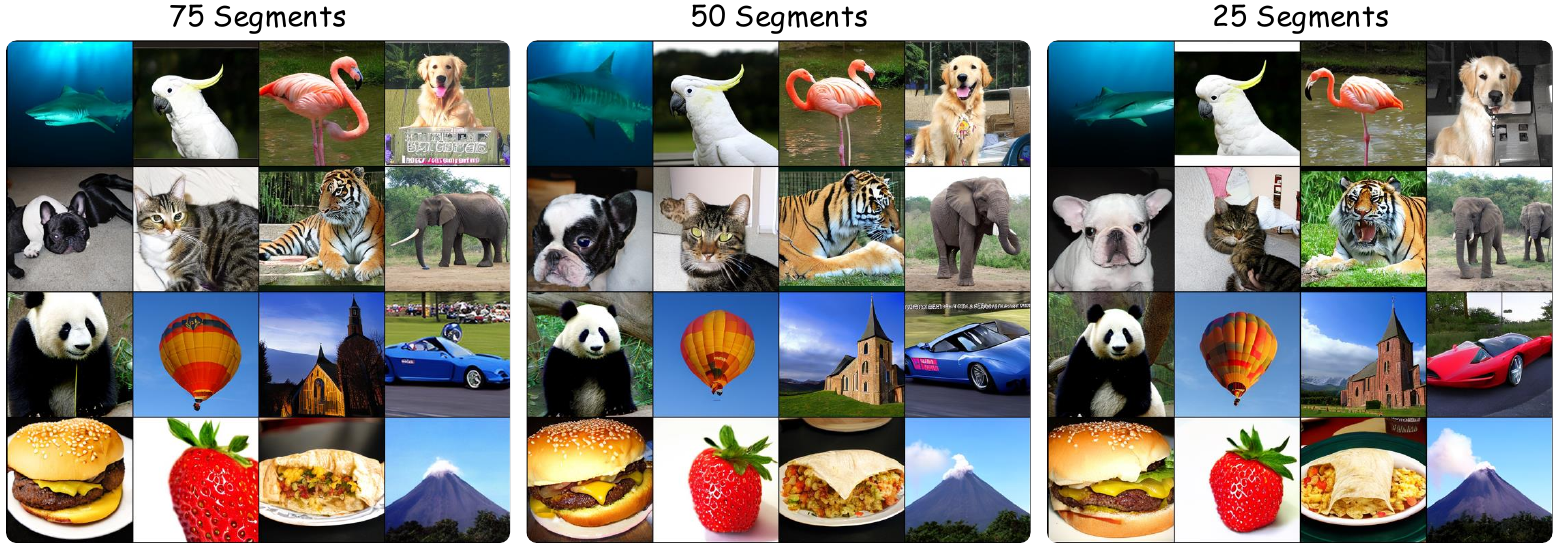}
    \caption{\textbf{Effect of segment count.} Varying the number of SLIC segments produces similar qualitative results, with 50 segments providing a balanced default.}
    \label{fig:appendix_segment_size_ablation}
\end{figure}

\begin{figure}[t]
    \centering
    \includegraphics[width=0.9\linewidth]{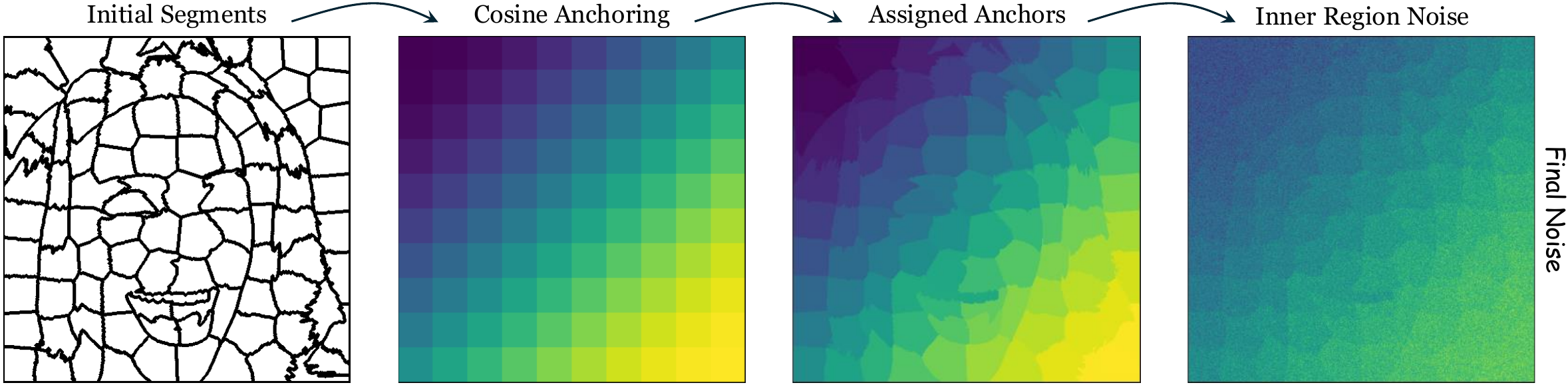}
    \caption{\textbf{Cosine-Induced Spatial Anchoring and Noise Construction.} Segment anchors are deterministically sampled from a fixed cosine grid, after which inner-region noise is added to construct the final structured source sample.}
    \label{fig:appendix_cosine_anchor}
\end{figure}

We conduct a small ablation to study the effect of the number of local regions used for
structured noise construction, controlled by the number of SLIC segments. Specifically, we
train three DiT models with 25, 50, and 75 segments, respectively, while keeping all other
training settings fixed. Each model is trained for 1.8M iterations.
Figure~\ref{fig:appendix_segment_size_ablation} shows qualitative samples from the final
checkpoints. Overall, we observe no major qualitative difference across these settings: all
three models produce reasonable samples and preserve the main benefits of structured noise.
However, the number of segments still provides a useful practical trade-off. Using too many
segments can make the source structure overly fine-grained, which may restrict the model and
reduce its ability to generate details that extend across segment boundaries. Conversely, using
too few segments weakens the spatial control provided by the structured source. Based on this
trade-off, we use 50 segments as a balanced default in our experiments.

\begin{figure}[t]
    \vspace{-6pt}
    \centering
    \includegraphics[width=0.9\linewidth]{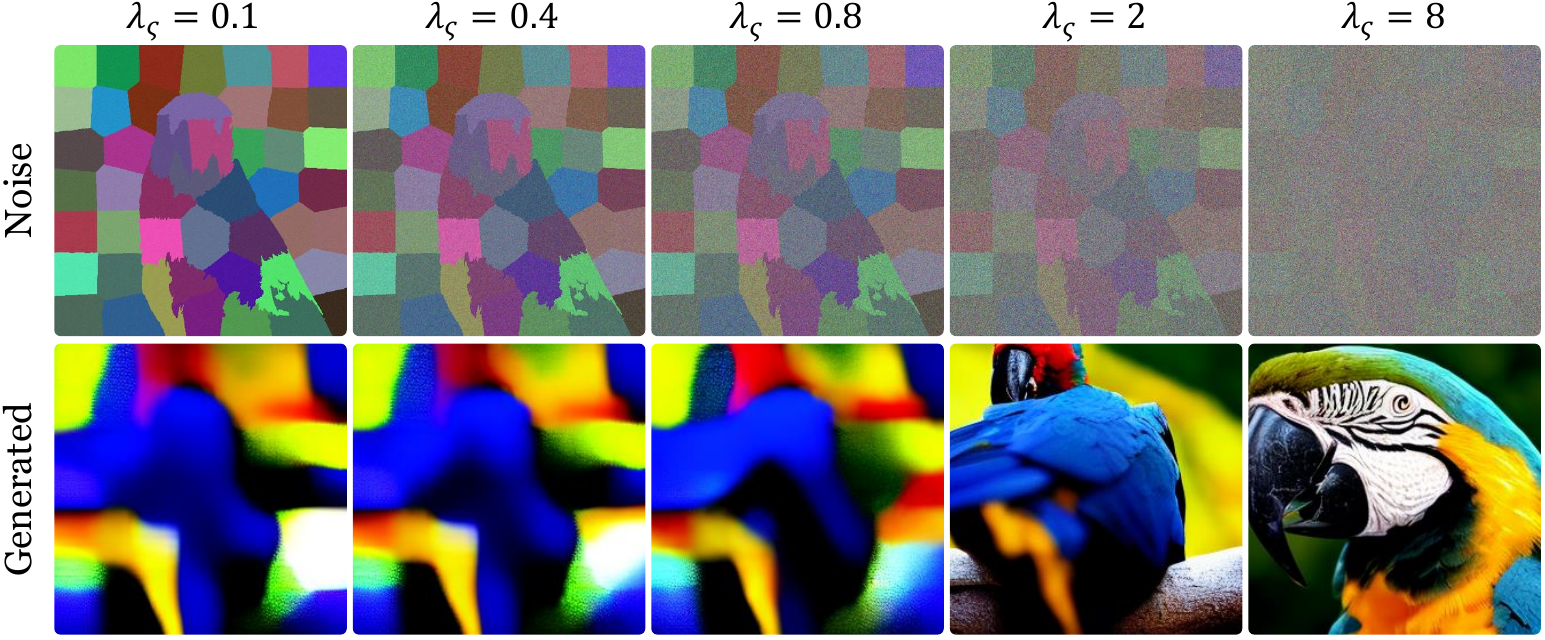}
    \caption{\textbf{Effect of $\lambda_\varsigma$.} Noise constructed with varying $\lambda_\varsigma$ and evaluated using a standard FM-based DiT. As $\lambda_\varsigma \rightarrow \infty$, the noise approaches i.i.d.\ Gaussian, matching the training distribution of the base model and yielding correct generation. This enables post-training via Progressive Coherence Annealing, by initializing with large $\lambda_\varsigma$ and gradually reducing it to introduce structure.}
    \label{fig:secondary_std_to_inf_visualization}
\end{figure}

We also explored replacing superpixels with semantic segmentation masks. In particular, for
FFHQ, we used face-part segmentations, such as eyes, hair, lips, and other facial regions, as
the local regions for structured noise construction. This setting was trainable and produced
reasonable samples. However, we found it less suitable for our goal for two main reasons.
First, semantic regions do not always correspond to local coherence: a large region such as
hair can cover spatially distant or visually diverse areas, where the pixels are not
necessarily strongly correlated. Second, semantic masks provide much coarser control than
superpixels, since each region often corresponds to an entire object part rather than a
fine-grained local area. 
construction.

\section{Limitations}
\label{sec:app:limitation}
While StructFlow introduces a new paradigm for source distribution design that improves localized controllability and structure preservation, it also has several limitations. First, our method relies on segmentation masks or mask proxies to construct the structured source distribution. Although we explore retrieval-based and training-free alternatives at inference time, generation quality can depend on how closely these masks match the distribution used during training. Second, the structured source introduces additional hyperparameters, such as the number of segments and the secondary noise scale $\lambda_{\varsigma}$, which control the strength of locality and may require tuning across datasets, resolutions, or model architectures.

% \newpage
% \input{checklist.tex}

\end{document}